%% file: top.tex
\documentclass[10pt,twocolumn,letterpaper]{article}

\usepackage{iccv}
\usepackage{times}
\usepackage{epsfig}
\usepackage{graphicx}
\usepackage{amsmath}
\usepackage{amssymb}
\usepackage{bm}
\usepackage{algorithm}
\usepackage{algpseudocode}

\usepackage{animate}

\makeatletter
\newcommand{\removelatexerror}{\let\@latex@error\@gobble}
\makeatother

\usepackage[pagebackref=true,breaklinks=true,letterpaper=true,colorlinks,bookmarks=false]{hyperref}

\iccvfinalcopy 

\def\iccvPaperID{3463} 

\ificcvfinal\fi
\begin{document}

\title{\vspace{-4mm}Meta-Sim: Learning to Generate Synthetic Datasets}

\author{
\vspace{1mm}
Amlan Kar$^{1,2,3}$
\hspace{0.5cm}
Aayush Prakash$^{1}$  
\hspace{0.5cm}
Ming-Yu Liu$^{1}$
\hspace{0.5cm}
Eric Cameracci$^{1}$
\hspace{0.5cm}
Justin Yuan$^{1}$
\\
\vspace{1em}
Matt Rusiniak$^{1}$
\hspace{0.5cm}
David Acuna$^{1,2,3}$
\hspace{0.5cm}
Antonio Torralba$^{4}$
\hspace{0.5cm}
Sanja Fidler$^{1,2,3}$\thanks{Correspondence to amlan@cs.toronto.edu, sfidler@nvidia.com}
\\
$^1$NVIDIA \hspace{2em} $^2$University of Toronto \hspace{2em}   $^3$Vector Institute  \hspace{2em} $^4$ MIT\\
}

\maketitle

\begin{abstract}
Training models to high-end performance requires availability of large labeled datasets, which are expensive to get. The goal of our work is to \emph{automatically synthesize} labeled datasets that are relevant for a downstream task. We propose Meta-Sim, which learns a generative model of synthetic scenes, and obtain images as well as its corresponding ground-truth via a graphics engine. We parametrize our dataset generator with a neural network, which learns to modify attributes of scene graphs obtained from probabilistic scene grammars, so as to minimize the distribution gap between its rendered outputs and target data. If the real dataset comes with a small labeled validation set, we additionally aim to optimize a meta-objective, \ie downstream task performance. Experiments show that the proposed method can greatly improve content generation quality over a human-engineered probabilistic scene grammar, both qualitatively and quantitatively as measured by performance on a downstream task. Webpage: \url{https://nv-tlabs.github.io/meta-sim/}
\end{abstract}

\input{intro}
\input{related}
\input{method}
\input{experiments}
\input{conclusion}
\input{supplementary}

{\small
\bibliographystyle{ieee}
\bibliography{references}
}

\end{document}

%% file: intro.tex
\vspace{-3mm}
\section{Introduction}
Data collection and labeling is a laborious, costly and time consuming venture, and represents a major bottleneck in most current machine learning pipelines. To this end, synthetic content generation~\cite{Sintel,Richter16,VirtualKITTI,sdr18} has emerged as a promising solution since all ground-truth comes for free -- via the graphics engine. It further enables us to train and test our models in virtual environments~\cite{RosCVPR16,Dosovitskiy17,House3D,THOR,AirSim} before deploying to the real world, which is crucial for both scalability and safety.
Unfortunately, an important performance issue arises due to the domain gap existing between the synthetic and real-world domains.

Addressing the domain gap issue has led to a plethora of work on synthetic-to-real domain adaptation~\cite{cycada,gradgan,pseudolabelyang,studentteachervisda,drjosh17,sdr18,tsai2018learning}. These techniques aim to learn domain-invariant features and thus more transferrable models. One of the mainstream approaches is to learn to stylize synthetic images to look more like those captured in the real-world~\cite{cycada,gradgan,cyclegan,unit,munit}. As such, these models address the \emph{appearance gap} between the synthetic and real-world domains. They share the assumption that the domain gap is due to the differences that are fairly low level.

Here, we argue that domain gap is also due to a \emph{content gap}, arising from the fact that the synthetic content (\eg layout and types of objects) mimics a limited set of scenes, not necessarily reflecting the diversity and distribution of objects of those captured in the real world. For example, the Virtual KITTI~\cite{VirtualKITTI} dataset was created by a group of engineers and artists,
to match object locations and poses in KITTI~\cite{kitti} which was recorded in Karlsruhe, Germany. But what if the target city changes to Tokyo, Japan, which has much heavier traffic and many more high-rise buildings? Moreover, what if the downstream task that we want to solve changes from object detection to lane estimation or rain drop removal? Creating synthetic worlds that ensure realism and diversity for any desired task requires significant effort by highly-qualified experts and does not scale to the fast demand of various commercial applications.

\begin{figure}[t!]
\vspace{-2mm}
\centering
\includegraphics[width=1.0\linewidth]{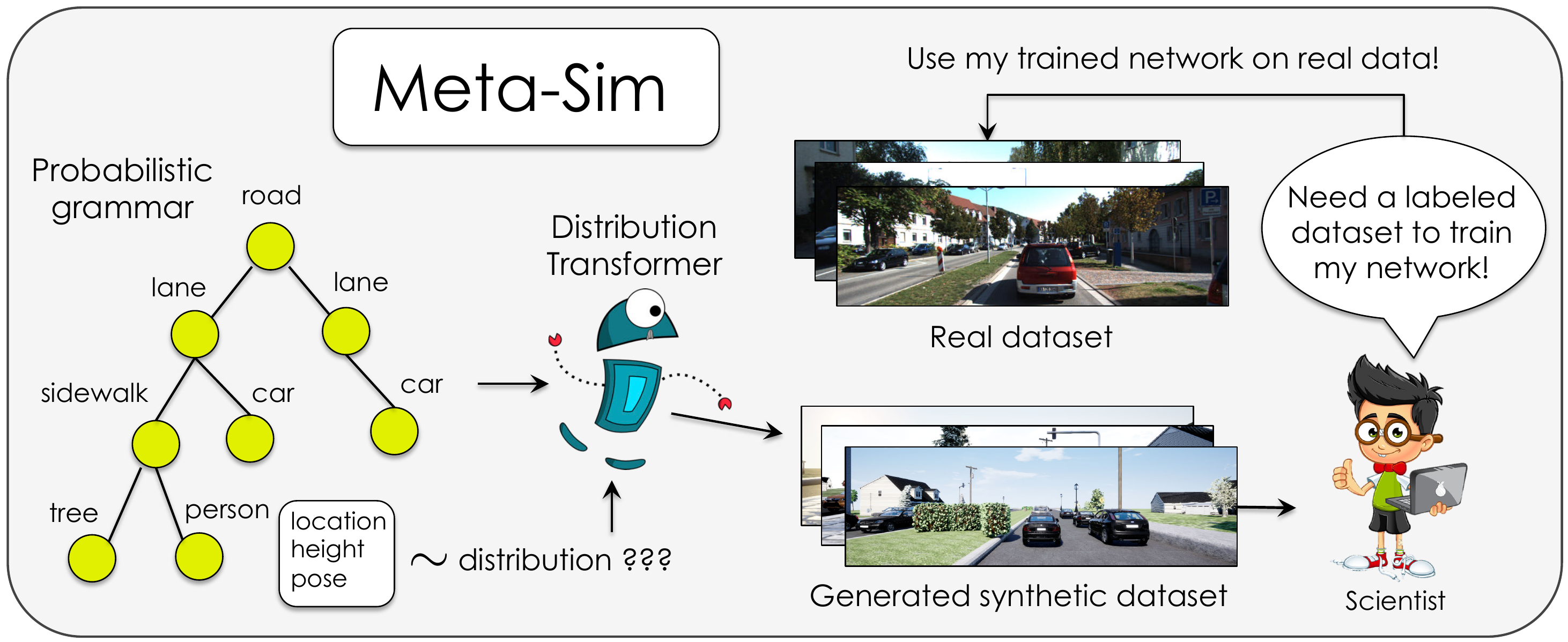}
\caption{Meta-Sim is a method to generate synthetic datasets that bridge the \emph{distribution gap} between real and synthetic data and are optimized for \emph{downstream task performance}}
\label{fig:model}
\vspace{2mm}
\end{figure}

In this paper, we aim to learn a generative model of synthetic scenes that, by exploiting a graphics engine, produces \emph{labeled} datasets with a content distribution matching that of imagery captured in the desired real-world datasets. Our \emph{Meta-Sim} builds on top of probabilistic scene grammars which are commonly used in gaming and graphics to create diverse and valid virtual environments. In particular, we assume that the structure of the scenes sampled from the grammar are correct (\eg a driving scene has a road and cars), and learn to modify their attributes. By modifying locations, poses and other attributes of objects, Meta-Sim gains a powerful flexibility of adapting scene generation to better match real-world scene distributions. 
\emph{Meta-Sim} also optimizes a meta objective of adapting the simulator to improve downstream real-world performance of a Task Network trained on the datasets synthesized by our model. Our learning framework optimizes several objectives  using approximated gradients through a non-differentiable renderer. 


We validate our approach on two toy simulators in controlled settings, where Meta-Sim is shown to excel at bridging the distribution gaps. We further showcase Meta-Sim on adapting a probabilistic grammar akin to SDR~\cite{sdr18} to better match a real self-driving dataset, leading to improved content generation quality, as measured by sim-to-real performance. To the best of our knowledge, Meta-Sim is the first approach to enable dataset and task specific synthetic content generation, and we hope that our work opens the door to more adaptable simulation in the future.

%% file: related.tex
\section{Related Work}
{\bf Synthetic Content Generation and Simulation.} The community has been investing significant effort in creating high-quality synthetic content, ranging from driving scenes~\cite{RosCVPR16,VirtualKITTI,Richter16,Dosovitskiy17,sdr18,synscapes}, indoor navigation~\cite{House3D}, household robotics~\cite{Virtualhome,THOR}, robotic control~\cite{Mujoco}, game playing~\cite{openaigym}, optical flow estimation~\cite{Sintel}, and quadcopter control and navigation~\cite{AirSim}.
While such environments are typically very realistic, they require qualified experts to spend a huge amount of time to create these virtual worlds. Domain Randomization (DR) is a cheaper alternative to such photo-realistic simulation environments~\cite{sadeghi2016cad2rl,drjosh17,sdr18}.
The DR technique generates a large amount of diverse scenes by inserting objects in random locations and poses. As a result, the distribution of the synthetic scenes is very different to that of the real world scenes. 
We, on the other hand, aim to align the synthetic and real distributions through a direct optimization on the attributes and through a meta objective of optimizing for performance on a down-stream task.

\textbf{Procedural modeling and probabilisic scene grammars} are an alternative approach to content generation\footnote{\tiny\url{https://www.sidefx.com/}}, which are able to produce worlds at the scale of full cities\footnote{\tiny\url{https://www.esri.com/en-us/arcgis/products/esri-cityengine/overview}}, and mimic diverse 3D scenes for self-driving\footnote{\tiny\url{https://www.paralleldomain.com/}}. However, the parameters for generating the distributions that control how a scene is generated need to be manually specified. This is not only tedious but also error-prone. There is no guarantee that the specified parameters can generate distributions that faithfully reflect real world distributions. ~\cite{kulkarni2015picture, mansinghka2013approximate} use such probabilistic programs to invert the generative process and infer a program given an image, while we aim to learn the generative process itself from real data.

{\bf Domain Adaptation} aims at addressing the domain gap, \ie the mismatch between the distribution of data used to train a model and the distribution of data that the model is expected to work with. From synthetic to real, two kinds of domain gaps arise: the appearance (style) gap and the content (layout) gap. Most existing work~\cite{cycada,gradgan,pseudolabelyang,studentteachervisda,cyclegan,unit,munit} tackle the former by using generative adversarial networks (GANs)~\cite{goodfellow2014generative} to transform the appearance distribution of the synthetic images to look more like that of the real images. Others~\cite{cycada,gradgan} add additional task based constraints to ensure that the layout of the stylized images remain the same. Other techniques use pseudo label based self learning~\cite{pseudolabelyang} and student-teacher networks~\cite{studentteachervisda} for domain adaptation. Our work is an early attempt to tackle the second kind of domain gap -- the content gap. We note that the appearance gap is orthogonal to the content gap, and prior art could be directly plugged into our method.
 
{\bf Optimizing Simulators.} Louppe \etal~\cite{louppe2017adversarial} attempt to optimize non-differentiable simulators with the key difference being in the method and the end goal. They optimize using a variational upperbound of a GAN-like objective to produce samples representative of a target distribution. We, on other hand, use the MMD~\cite{gretton2012kernel} distance metric for comparing distributions and also optimize a meta objective to produce samples suitable for a downstream task. ~\cite{chebotar2018closing} learn to optimize simulator parameters for robotic control tasks, where trajectories between the real and simulated robot can be directly compared. ~\cite{ruiz2018learning} optimize high level exposed parameters by optimizing for downstream task performance using Reinforcement Learning. We, on the other hand, optimize low level scene parameters (at the level of every object) while also learning to match distributions along with optimizing downstream task performance. Ganin \etal~\cite{ganin2018synthesizing} attempt to synthesize images by learning to generate even lower-level programs (at the level of brush strokes) that a graphics engine can interpret to generate realistic looking images, as measured by a trained discriminator.

%% file: method.tex
\section{Meta-Sim}

In this section, we introduce \emph{Meta-Sim}. Given a dataset of real imagery $X_R$ and a task $T$ (\eg object detection), our goal is to \emph{synthesize} a training dataset $D_T=(X_T,Y_T)$ with $X_T$ synthesized imagery that resembles the given real imagery, and $Y_T$ the corresponding ground-truth for task $T$. To simplify notation, we omit subscript $T$ from here on. 

We parametrize data synthesis with a neural network, \ie $D(\theta)=(X(\theta), Y(\theta))$. Our goal in this paper is to learn the parameters $\theta$ such that the distribution of $X(\theta)$ matches that of $X_R$ (real imagery). Optionally, if the real dataset comes with a small validation set $V$ that is labeled for task $T$, we additionally aim to optimize a meta-objective, \ie downstream task performance. The latter assumes we also have a trainable task solving module (\ie another neural network), the performance of which we want to maximize by training it on our generated training data. We refer to this module as a Task Network, which will be treated as a black box in our work. 
Note that Meta-Sim has parallels to Neural Architecture Search~\cite{zoph2016neural}, where our search is over the input datasets to a fixed neural network instead of a search over the neural network architecture given fixed data.

\textbf{Image Synthesis vs Rendering.} Generative models of pixels have only recently seen success in generating realistic high resolution images~\cite{brock2018large, karras2018style}. Extracting task specific ground-truth (eg: segmentation) from them remains a challenge. Conditional generative models of pixels condition on input images and transform their appearance, producing compelling results. However, these methods assume ground truth labels remain unchanged, and thus are limited in their \emph{content} (structural) variability. 
In Meta-Sim we aim to \emph{learn} a {\bf generative model of synthetic 3D content}, and obtain $D$ via a graphics engine. Since the 3D assets come with semantic information (\ie, we know an asset is a \emph{car}), compositing or modifying the synthetic scenes will still render perfect ground-truth. The main challenge is to learn the 3D scene composition by optimizing solely the distribution mismatch of rendered with real imagery.
The following subsections layout Meta-Sim in detail and are structured as follows: Sec.~\ref{ss:scene_graph} introduces the representation of parametrized synthetic worlds,
while Sec.~\ref{ss:training} describes our learning framework. 

\subsection{Parametrizing Synthetic Scenes}
\label{ss:scene_graph}

\begin{figure}
\vspace{-2mm}
\centering
\includegraphics[width=1.0\linewidth]{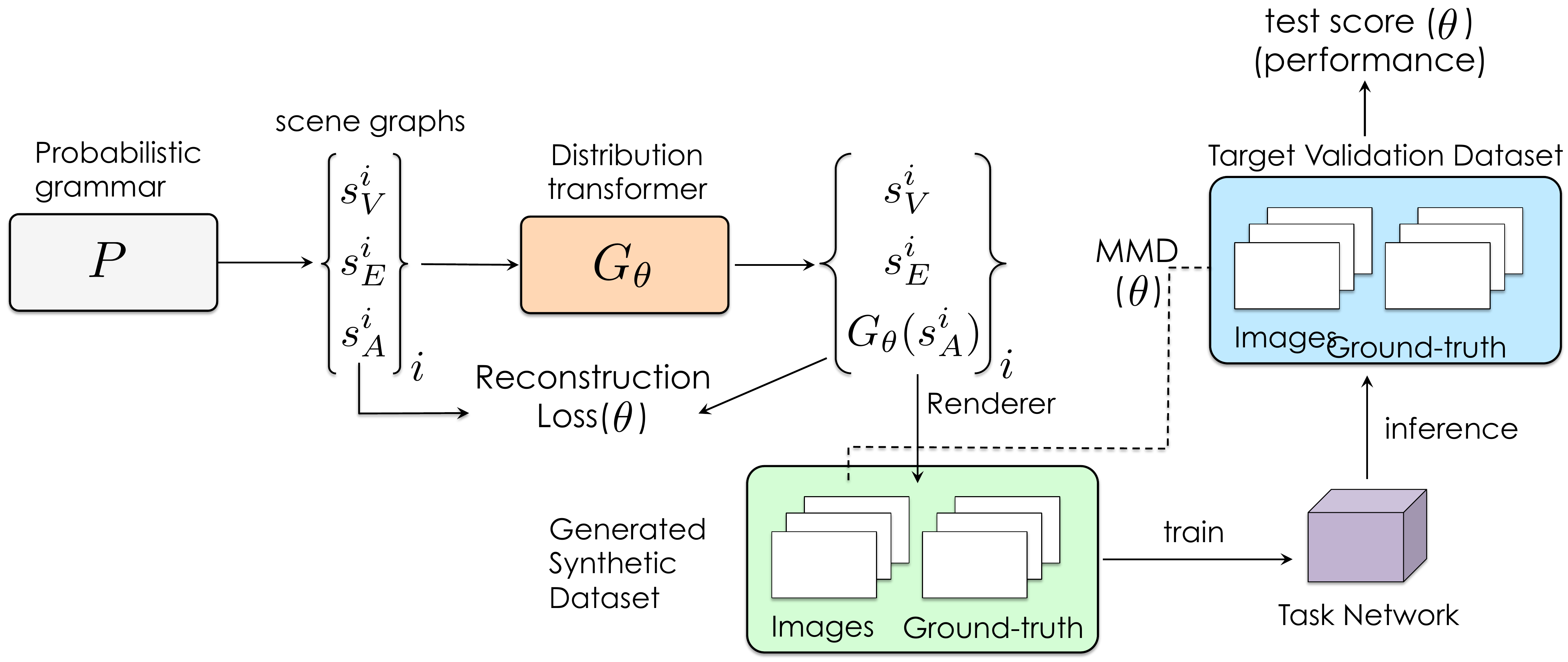}
\caption{Overview of Meta-Sim: The goal is to learn to transform samples coming from a probabilistic grammar with a distribution transformer, aiming to minimize the \emph{distribution gap} between simulated and real data and maximize \emph{sim-to-real performance}}
\label{fig:model}
\vspace{2mm}
\end{figure}

\begin{figure*}
\centering
\begin{minipage}{0.22\textwidth}
\centering
\includegraphics[width=0.99\linewidth,trim=0 138 330 0,clip]{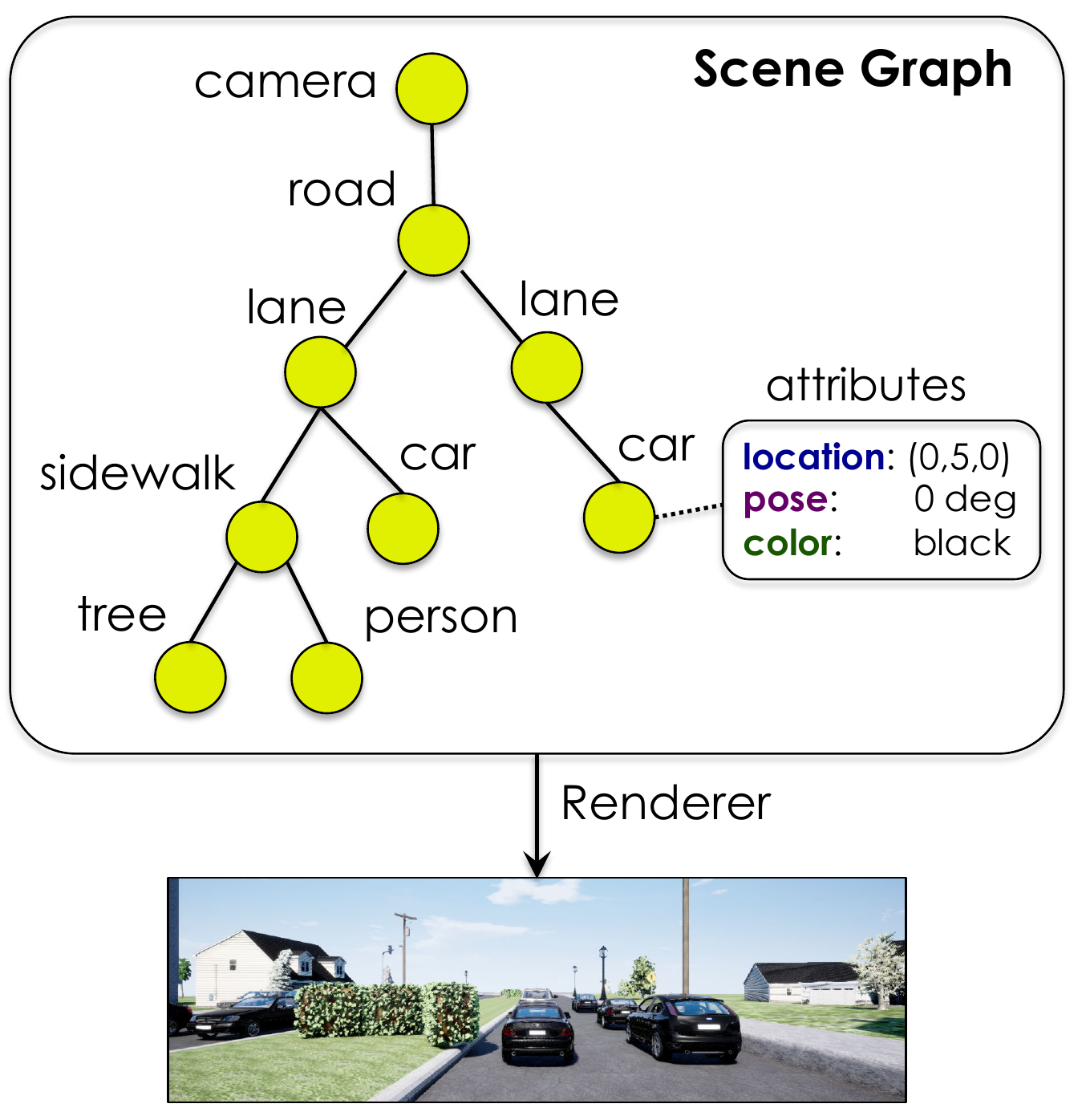}
\caption{\small Simple scene graph example for a driving scene.}
\label{fig:scenegraph}
\end{minipage}
\hspace{2mm}
\begin{minipage}{0.758\textwidth}
\centering
\includegraphics[width=\linewidth]{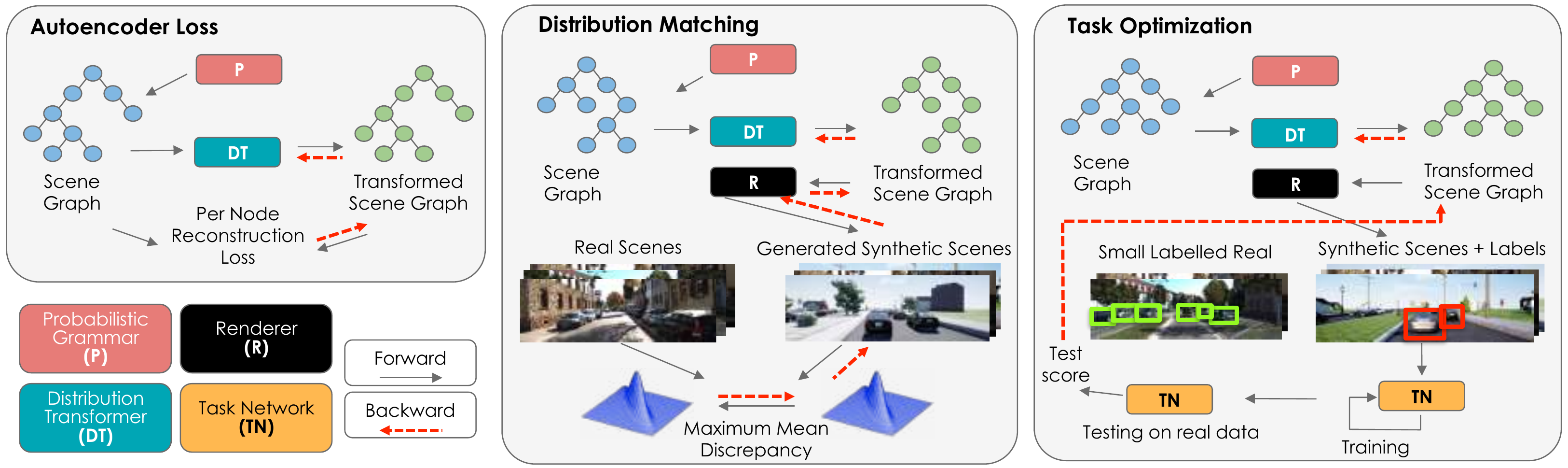}
\caption{\small Illustration of different losses used in \emph{Meta-Sim}, including forward and backward pass control flow for each step. 
We indicate transformed attributes of a scene graph by changing colors of the nodes.}
\label{fig:training}
\end{minipage}
\vspace{-4mm}
\end{figure*}

\paragraph{Scene Graphs} are a common way to represent 3D worlds in gaming/graphics. A scene graph represent elements of a scene in a concise hierarchical structure, with each element having a set of attributes (eg. class, location, or even the id of a 3D asset from a library) (see Fig.~\ref{fig:scenegraph}). The hierarchy defines parent-child dependencies, where the attributes of the child elements are typically defined relative to the parent's, allowing for an efficient and natural way to create and modify scenes. 
The corresponding image and pixel-level annotations can be rendered easily by placing objects as described in the scene graph. 

In order to generate \emph{diverse} and \emph{valid} 3D worlds, the typical approach is to specify the generative process of the graph by a \emph{probabilistic scene grammar}~\cite{zhu2007stochastic}. 
For example, to generate a traffic scene, one might first lay out the centerline of the road, add parallel lanes, position aligned cars on each lane, etc. The structure of the scene is defined by the grammar, while the attributes are typically sampled from parametric distributions, which require careful tuning.

In our work, we assume access to a probabilistic grammar from which we can sample initial scene graphs. We assume the \emph{structure} of each scene graph is correct, \ie the driving scene has a road, sky, and a number of objects. This is a reasonable assumption, given that inferring structure (inverse graphics) is known to be a hard problem. Our goal is to modify the \emph{attributes} of each scene graph, such that the transformed scenes, when rendered, will resemble the distribution of the real scenes. By modifying the attributes, we give the model a powerful flexibility to change objects' locations, poses, colors, asset ids, etc. This amounts to learning a conditional generative model, which, by conditioning on an input scene graph  transforms its node attributes. In essence, we keep the \emph{structure} generated by the probabilistic grammar, but transform the distribution of the \emph{attributes}. 
Thus, our model acts as a \emph{Distribution Transformer}. 

\vspace{-4mm}
\paragraph{Notation.} Let  $P$ denote the probabilistic grammar from which we can sample scene graphs $s \sim P$. 
We denote a single scene graph $s$ as a set of vertices $s_V$, edges $s_E$ and attributes $s_A$. We have access to a renderer $R$, that can take in a scene graph $s$ and generate the corresponding image and ground truth, $R(s) = (x,y)$. 
Let $G_\theta$ refer to our Distribution Transformer, which takes an input scene graph $s$ and outputs a scene graph $G_\theta(s)$, with transformed attributes but the same structure, \ie  $G_\theta(s=[s_V, s_E, s_A])$ = $[s_V, s_E, G_\theta(s_A)]$. Note that by sampling many scene graphs, transforming their attributes, and rendering, we obtain a synthetic dataset $D(\theta)$.

\vspace{-4mm}
\paragraph{Architecture of $G_\theta$.} Given the graphical structure of scene graphs, modeling $G_\theta$ via a Graph Neural Network is a natural choice. In particular, we use Graph Convolutional Networks (GCNs)~\cite{kipf2016semi}. We follow~\cite{yao2018exploring} and use a graph convolutional layer that utilizes two different weight matrices to capture top-down and bottom-up information flow separately. Our model makes per node predictions \ie generates transformed attributes $G_\theta(s_A)$ for each node in $s_V$. 

\vspace{-4mm}
\paragraph{Mutable Attributes:} We input to $G_\theta$ all attributes $s_A$, but we might want to only modify specific attributes and trust the probabilistic grammar $P$ on the rest.
For example, in Fig.~\ref{fig:scenegraph} we may not want to change the heights of houses, or width of the sidewalks, if our final task is car detection. This reduces the number of exposed parameters our model is tasked to tune thus improving training time and complexity. 
Therefore, in the subsequent parts, we assume we have a subset of attributes per node $v \in s_V$ which are mutable (modifiable), denoted by $s_{A,mut}(v)$. From here onwards, it is assumed that only the mutable attributes in $s_{A,mut}(v) \forall v$ are changed by $G_\theta$; others remain the same as in $s$. 

\vspace{-1mm}
\subsection{Training Meta-Sim}
\label{ss:training}

We now introduce our learning framework. Since our learning problem is very hard and computationally intensive, we first pre-train our model using a simple autoencoder loss in Sec.~\ref{sss:recons}. The distribution matching loss is presented in Sec~\ref{sss:distmatch}, while meta-training is described in Sec~\ref{sss:task}. The overview of our model is given in Fig.~\ref{fig:model}, with the particular training objectives illustrated in Fig.~\ref{fig:training}.

\vspace{-4mm}
\subsubsection{Pre-training: Autoencoder Loss} 
\label{sss:recons}
\vspace{-2mm}
A probabilistic scene grammar $P$ represents a prior on how a scene should be generated. Learning this prior is a natural way to pre-train our \emph{Distribution Transformer}. This amounts to training $G_\theta$ to perform the identity function \ie $G_\theta(s) = s$. The input feature of each node is its attribute set ($s_A$), which is defined consistently across all nodes (see suppl.). Since $s_A$ is composed of different categorical and continuous components, appropriate losses are used per feature component when training to reconstruct (\ie cross-entropy loss for categorical attributes, and L1 loss for continuous attributes). 

\vspace{-4mm}
\subsubsection{Distribution Matching}
\label{sss:distmatch}
\vspace{-2mm}
The first objective of training our model is to bring the distribution of the rendered images to be closer to the distribution of real imagery $X_R$. The Maximum Mean Discrepancy (MMD)~\cite{gretton2012kernel} metric is a frequentist measure of the similarity of two distributions and has been used for training generative models~\cite{dziugaite2015training, li2015generative, li2017mmd} to match statistics of the generated distribution with the target distribution. An alternative, adversarial learning with discriminators, however, is known to suffer from mode collapse, and a general instability in training. Pixel-wise generative models with MMD have usually suffered from not being able to model high-frequency signals (resulting in blurry generations). Since our generative process goes through a renderer, we sidestep the issue altogether, and thus choose MMD for training stability. 

We compute MMD in the feature space of an InceptionV3~\cite{szegedy2016rethinking} network (known as Kernel Inception Distance (KID)~\cite{binkowski2018demystifying}). This feature extractor is denoted by the function $\phi$. We use the kernel trick for the computation with a gaussian kernel $k(x_i, x_j)$. We refer the reader to~\cite{li2015generative} for more details. The \emph{Distribution Matching} box in Fig.~\ref{fig:training} shows the training procedure pictorially. Specifically, given scene graphs $s_1, ..., s_N$ sampled from $P$ and target real images $X_R$, the squared MMD distance can be computed as,
\begin{small}
\vspace{-4mm}
\begin{align}
\label{eq:mmd}
\mathcal{L}_{{MMD}^2} &= \frac{1}{N^2}\sum_{i=1}^{N}\sum_{i'=1}^N k(\phi(X_\theta(s_i)), \phi(X_\theta(s_{i'}))\nonumber\\
&+\frac{1}{M^2} \sum_{j=1}^{M}\sum_{j'=1}^{M} k(\phi(X_R^j), \phi(X_R^{j'}))\nonumber\\
&-\frac{1}{MN} \sum_{i=1}^{N}\sum_{j=1}^{M} k(\phi(X_\theta(s_i)), \phi(X_R^{j}))
\end{align}
\vspace{-2mm}
\end{small}
where the image rendered from $s$ is $X_\theta(s)=R(G_\theta(s)))$.
\vspace{-2mm}
\paragraph{Sampling from $\bm{G_\theta(s)}$.} For simplicity, we overloaded the notation $R(G_\theta(s))$, since $R$ would actually require sampling from the prediction $G_\theta(s)$. In general, we assume independence across scenes, nodes and attributes, which lets each attribute of each node in the scene graph be sampled independently. While training with $MMD$, we mark categorical attributes in $s_A$ as immutable. The predicted continuous attributes are directly passed as the sample.
\vspace{-2mm}
{
\begin{algorithm}[t!]
\begin{footnotesize}
\begin{algorithmic}[1]
\State \textbf{Given:} $P, R, G_\theta$ \Comment{{\scriptsize Probabilistic grammar, Renderer, GCN Model}}
\State \textbf{Given:} $\texttt{TaskNet}, X_R, V$ \Comment{{\scriptsize Task Model, Real Images, Target Validation Data}}
\State \textbf{Hyperparameters:} $E_m, I_m, B_m$ \Comment{{\scriptsize Epochs, Iters, Batch size}}
\While{$e_m \le E_m$} \Comment{{\scriptsize Meta training}}
\State $loss$ = 0;\;
\State $data$ = []; $samples$ = []; \Comment{{\scriptsize Caching data \& samples generated in epoch}}
	\While{$i_m \le I_m$}
	\State $S$ = $G_\theta$(sample($P$, $B_m$)); \Comment{{\scriptsize Generate $B_m$ samples from $P$}}
	\State \hfill{\scriptsize and transform them}
	\State $D$ = $R(S)$; \Comment{{\scriptsize Render images, labels from $S$}}
	\State $data$ += $D$; $samples$ += $S$;\;
	\State $loss$ += $\mathcal{L}_{{MMD}^2}(D, X_R)$; \Comment{{\scriptsize MMD between generated and}} 
	\State \hfill{{\scriptsize target real images}} 
	\EndWhile
	\State $\texttt{TaskNet}$ = train($\texttt{TaskNet}$, $data$);  \Comment{{\scriptsize Train \texttt{TaskNet} on \emph{data}}} 
	\State $score$ = test($\texttt{TaskNet}$, $V$);   \Comment{{\scriptsize Test \texttt{TaskNet} on target val}} 
	\State $loss$ += $- (score - moving\_avg(score))\cdot \log p_{G_\theta}(samples)$ \Comment{{\scriptsize Eq.~\ref{eq:taskloss}}}
	\State $G_\theta$ = optimize($G_\theta$, $loss$); \Comment{{\scriptsize SGD step}}
\EndWhile
\end{algorithmic}
\end{footnotesize}
\caption{\small{Pseudocode for Meta-Sim's meta training phase}}
\label{algo:train}
\end{algorithm}
}

\vspace{-3mm}
\paragraph{Backprop through a Renderer.} 
\label{sss:diffrender}
For optimizing the MMD loss, 
we need to backpropagate the gradient through the non-differentiable rendering function $R$. 
The gradient of $R(G_\theta(s))$ w.r.t. $G_\theta(s)$ can be approximated using the method of finite differences\footnote{computed by perturbing each attribute in the scene graph $G_\theta(s)$}. While this gives us noisy gradients, we found it sufficient to be able to train our models in practice, with the benefit of being able to use photorealistic rendering. We note that recent work on differentiable rendering~\cite{kato2018neural,Li2018DMC} could potentially benefit this work. 

\vspace{-2mm}
\subsubsection{Optimizing Task Performance}
\label{sss:task}
\vspace{-2mm}
The second objective of training the model $G_\theta$ is to generate data $R(G_\theta(S))$ given samples $S = \{s_1, ..., s_K\}$ from the probabilistic grammar $P$, such that a model trained on this data achieves best performance when tested on target data $V$. This can be interpreted as a meta-objective, where the input data must be optimized to improve accuracy on a validation set. We introduce a task network $\texttt{TaskNet}$ to train using our data and to measure validation performance on. We train $G_\theta$ under the following objective,
\begin{small}
\vspace{-2mm}
\begin{align}
\max_\theta \quad &\mathbb{E}_{S' \sim G_\theta(S)} \big{[}\texttt{score}(S')\big{]}
\label{eq:tasknet}
\end{align}
\end{small}
where $\texttt{score}(S')$ is the performance metric achieved on validation data $V$ after training $\texttt{TaskNet}$ on data $R(G_\theta(S'))$. The task loss in Eq.~\ref{eq:tasknet} is not differentiable w.r.t the parameters $\theta$, since the score is measured using validation data and not $S'$. We use the REINFORCE score function estimator (which is an unbiased estimator of the gradient) to compute the gradients of Eq.~\ref{eq:tasknet}. Reformulating the objective as a loss and writing the gradient gives,
\begin{small}
\begin{align}
\label{eq:taskloss}
\mathcal{L}_{task} &= -\mathbb{E}_{S' \sim G_\theta(S)} \big{[} \texttt{score}(S')\big{]}\\
\nabla_\theta \mathcal{L}_{task} &= -\mathbb{E}_{S' \sim G_\theta(S)}\big{[} \texttt{score}(S') \times \nabla_\theta \log p_{G_\theta}(S')\big{]}\nonumber
\end{align}
\end{small}
To reduce the variance of the gradient from the estimator above, we keep track of an exponential moving average of previous scores and subtract it from the current score~\cite{greensmith2004variance}. We approximate the expectation using one sample from $G_\theta(S)$. The \emph{Task Optimization} box in Fig.~\ref{fig:training} provides a pictorial overview of the task optimization.

\vspace{-4mm}
\paragraph{Sampling from $\rm{G_\theta(s)}$.} Eq.~\ref{eq:taskloss} requires us to be able to sample (and measure its likelihood) from our model. For continuous attributes, we interpret our model to be predicting the mean of a normal distribution per attribute, with a pre-defined variance. We use the reparametrization trick to sample from this normal distribution. For categorical attributes, it is possible to sample from a multinomial distribution from the predicted log probabilities per category. In this paper, we keep categorical attributes immutable.

\vspace{-4mm}
\paragraph{Calculating $\rm{\log p_{G_\theta}(S')}$.} Since we assume independence across scenes, attributes and objects in the scene, the likelihood in Eq~\ref{eq:taskloss} for the full scene is simply factorizable,
\begin{small}
\begin{equation}
\label{eq:factor}
\log p_G(S') = \sum_{s' \in S'} \sum_{v \in s'_{V}} \sum_{a \in s'_{A,mut}(v)} \log p_{G_\theta}(s'(v,a))
\end{equation}
\end{small}
where $s'(v,a)$ represents the attribute $a$ at node $v$ in a single scene $s'$ in batch $S'$. Note that the sum is only over mutable attributes per node $s_{A,mut}(v)$. The individual log probabilities come from the defined sampling procedure.
\vspace{-4mm}
\paragraph{Training Algorithm.} The algorithm for training with Distribution Matching and Task Optimization is presented in Algorithm~\ref{algo:train}.
\vspace{-2mm}

%% file: experiments.tex
\section{Experiments}
We evaluate Meta-Sim on three target datasets with three different tasks. The subsequent sections follow a general structure where we first outline the desired task, the target data and the task network\footnote{Task Network training details in suppl. material}. Then, we describe the probabilistic grammar 
that the \emph{Distribution Transformer} utilizes for its input, and the associated renderer that generates labeled synthetic data. We show quantitative and qualitative results after training the task network using synthetic data generated by Meta-Sim. We show strong boosts in quantitative performance and noticeable qualitative improvements in content-generation quality.

The first two experiments presented are in a controlled  setting, each with increasing complexity. The aim here is to probe Meta-Sim's capabilities when the shift between the target data distribution and the input distribution is known. The input distribution refers to the distribution of the scenes generated by samples from the probabilistic grammar that our \emph{Distribution Transformer} takes as input. Target data for these tasks is created by carefully modifying the parameters of the probabilistic program, which represents a known distribution gap that the model must learn.

\subsection{MNIST}
We first evaluate our approach on digit \textbf{classification} on MNIST-like data. The probabilistic grammar samples a background texture, one digit texture (image) from the MNIST dataset~\cite{lecun1998mnist} (which has an equal probability for any digit), and then samples a rotation and location for the digit. The renderer transforms the texture based on the sampled transformation and pastes it onto a canvas. 
\vspace{-4mm}
\paragraph{Task Network.} Our task network is a small $2$-layer CNN followed by $3$ fully connected layers. We apply dropout in the fully connected layers (with 50, 100 and 10 features). We verify that this network can achieve greater than $99\%$ accuracy on the regular MNIST classification task. We do not use data-augmentation while training (in all following experiments as well), as it might interfere with our model's training by changing the configuration of the generated data, making the task optimization signal unreliable. 

\begin{figure}[t!]
\vspace{-2mm}
\centering
\includegraphics[width=1\columnwidth]{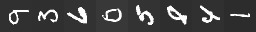}
\vspace{-4mm}
\caption{Examples from the rotated-MNIST dataset}
\label{im:rot}
\vspace{-4mm}
\end{figure}

\begin{figure}[t!]
\vspace{-2mm}
\centering
\includegraphics[width=1\columnwidth]{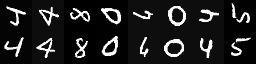}
\caption{\small{\textbf{(bottom)} Input scenes, \textbf{(top)} Meta-Sim's generated examples for MNIST with rotation gap}}
\label{fig:mnist1}
\vspace{1mm}
\centering
\includegraphics[width=1\columnwidth]{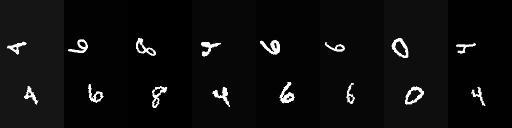}
\caption{\small{\textbf{(bottom)} Input scenes, \textbf{(top)} Meta-Sim's generated examples for MNIST with rotation and translation gap}}
\label{fig:mnist2}
\end{figure}

\paragraph{Rotating MNIST.}
In our first experiment, the probabilistic grammar generates input samples that are upright and centered, like regular MNIST digits (Fig~\ref{fig:mnist1} bottom). The target data $V$ and $X_R$ are images (at $32\times32$ resolution) where digits centered and always rotated by 90 degrees (Fig~\ref{im:rot}). Ideally, the model will learn this exact transformation, and rotate the digits in the input scene graph while keeping them in the same centered position.

\begin{figure}[t!]
\centering
\includegraphics[width=1\columnwidth]{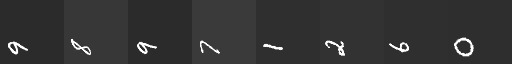}
\vspace{-4mm}
\caption{Examples from the rotated and translated MNIST}
\label{im:rot_trans}
\vspace{0.5mm}
\end{figure}

\vspace{-4mm}
\paragraph{Rotating and Translating MNIST.}
For the second experiment, we additionally add translation to the distribution gap, making the task harder for Meta-Sim. We generate $V$ and $X_R$ as 1000 images (at $64\times64$ resolution) where in addition to being rotated by 90 degrees, the digits are moved to the bottom left corner of the canvas (Fig~\ref{im:rot_trans}). The input probabilistic grammar remains the same, \ie one that generates centered and upright digits (Fig.~\ref{fig:mnist2} bottom).

\vspace{-4mm}
\paragraph{Quantitative Results.}
Table~\ref{tbl:mnist} shows classification on the target datasets with the two distribution gaps described above. The target datasets are fresh samples from the target distribution (separate from $V$). Training directly on the input scenes (coming from the input probabilistic grammar \ie generating upright and centered digits in this case) results in just above random performance. Our model recovers the transformation causing the distribution gap, and achieves greater than 99\% classification accuracy.

\begin{table}[h!]
\vspace{-2mm}
\centering
\begin{small}
\begin{tabular}{c|c|c}
 Data & Rotation & Rotation + Translation\\
 \hline\hline
 Prob. Grammar & 14.8 & 13.1\\
 Meta-Sim & \textbf{99.5} & \textbf{99.3}\\
 \hline
\end{tabular}
\end{small}
\vspace{1mm}
\caption{Classification performance on our MNIST with different distribution gaps in the data}
\label{tbl:mnist}
\vspace{-2mm}
\end{table}

\vspace{-4mm}
\paragraph{Qualitative Results.}
Fig.~\ref{fig:mnist1} and Fig.~\ref{fig:mnist2} show generations from our model at the end of training, and compares with the input scenes. Clearly, the model has learnt to perfectly transform the input distribution to replicate the target distribution, corroborating our quantitative results.

\vspace{-2mm}
\subsection{Aerial Views (2D)}
Next, we evaluate our approach on \textbf{semantic segmentation} of simulated aerial views of simple roadways. In the probabilistic grammar, we sample a background grass texture, followed by a (straight) road at some location and rotation on the background. Next, we sample two cars with independent locations (constrained to be in the road by parametrizing in the road's coordinate system), and rotations. In addition, we also sample a tree and a house randomly in the scene. Each object in the scene gets a random texture from a set of textures we collected for each object. We ended up with nearly 600 car, 40 tree, 20 house, 7 grass and 4 road textures. Overall, this grammar has more complexity than MNIST, due to the scene graphs having higher depth, more objects, and variability in appearance.

\begin{figure}[t!]
\vspace{-2mm}
\centering
\includegraphics[width=1\columnwidth]{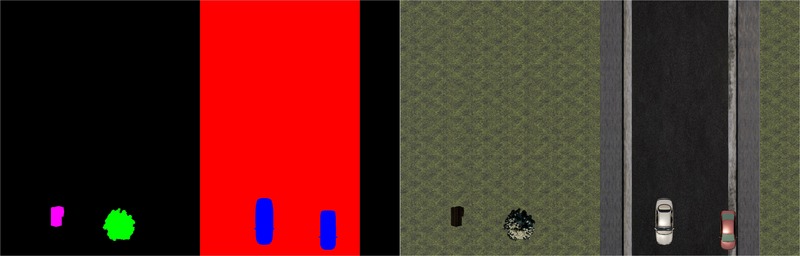}
\vspace{-4mm}
\caption{\small Example label and image from Aerial2D validation}
\label{fig:aerialval}
\vspace{-4mm}
\end{figure}

\begin{figure}[t!]
\centering
\includegraphics[width=1\columnwidth]{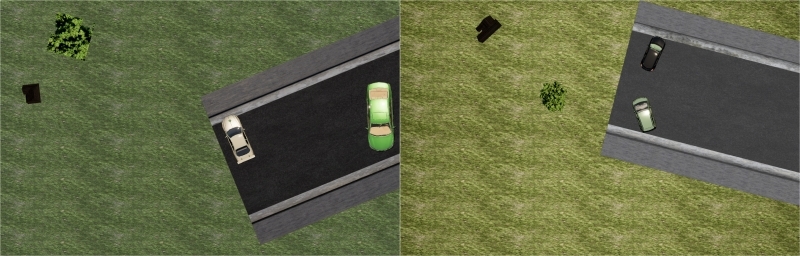}
\caption{\small{Example input scenes for Aerial2D}}
\label{fig:aerialtrain}
\end{figure}

$V$ and $X_R$ are created by tuning the grammar parameters to generate a realistic aerial view. (Fig.~\ref{fig:aerialval}). The input probabilistic grammar uses random parameters (Fig.~\ref{fig:aerialtrain}) bottom.

\vspace{-4mm}
\paragraph{Task Network.} We use a small U-Net architecture~\cite{ronneberger2015u} with a total of 7 convolutional layers (with $16$ to $64$ filters in the convolution layers) as our task-network.

\vspace{-4mm}
\paragraph{Quantitative Results.}
Table~\ref{tbl:aerialquant} shows semantic segmentation results on the target set. The results show that Meta-Sim effectively transforms the outputs of the probabilistic grammar, even in this relatively more complex setup, and improves the mean IoU. Specifically, it learns to drastically reduce the gap in performance for cars and also improves performance on roads.

\begin{figure}[t!]
\vspace{-2mm}
\centering
\includegraphics[width=1\columnwidth]{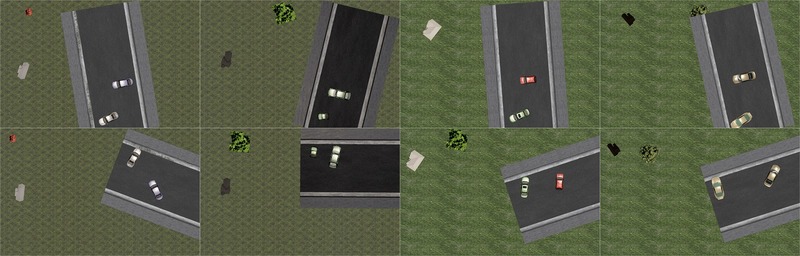}
\caption{\footnotesize {\bf (bottom)} input scenes, {\bf (top)} Meta-Sim's generated examples for Aerial semantic segmentation}
\label{fig:2dpairs}
\end{figure}

\vspace{-4mm}
\paragraph{Qualitative Results.} Qualitative results in Fig.~\ref{fig:2dpairs} show that the model indeed learns to exploit the convolutional structure of the task network, by only learning to orient. This is sufficient to achieve its job since convolutions are translation equivariant, but not rotation equivariant. 

\begin{table}[t!]
\vspace{-4mm}
\centering
\begin{small}
\begin{tabular}{c|c|c|c|c|c}
Data & Car & Road & House & Tree & Mean\\
\hline\hline
Prob. Grammar & 30.0 & 93.1 & \textbf{98.3} & \textbf{99.7} & 80.3\\
MetaSim & \textbf{86.7} & \textbf{99.6} & 95.0 & 99.5 & \textbf{95.2}\\
\hline
\end{tabular}
\end{small}
\vspace{1mm}
\caption{Semantic segmentation results (IoU) on Aerial2D}
\label{tbl:aerialquant}
\end{table}

\subsection{Driving Scenes (3D)}
After validating our approach on controlled experiments in a simulated setting, we now evaluate our approach for \textbf{object detection} on the challenging KITTI~\cite{kitti} dataset. KITTI was captured with a camera mounted on top of a car driving around the city of Karlsruhe in Germany. It consists of challenging traffic scenarios and scenes ranging from highways to urban to more rural neighborhoods. Contrary to the previous experiments, the distribution gap which we wish to reduce arises naturally here.

\begin{figure*}
\vspace{-3mm}
    \centering
    \addtolength{\tabcolsep}{-4.6pt}
    \begin{tabular}{ccc}
\includegraphics[width=0.32\textwidth,trim=0 0 0 40,clip]{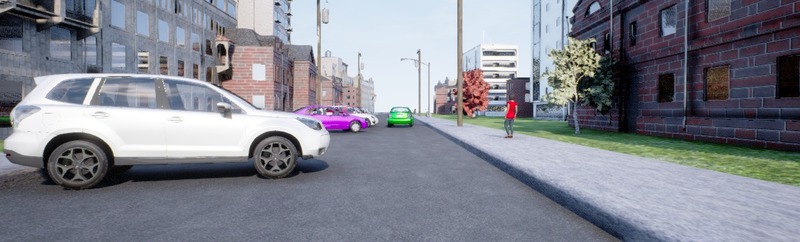} &
\includegraphics[width=0.32\textwidth,trim=0 0 0 40,clip]{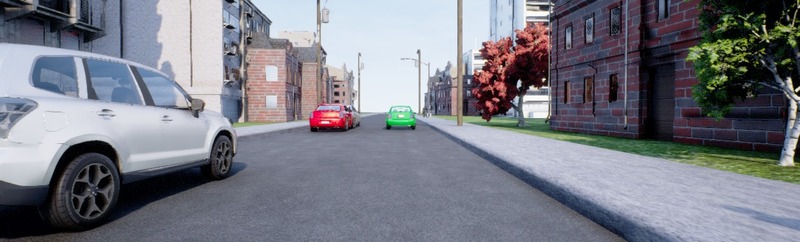} & 			\includegraphics[width=0.32\textwidth,trim=0 0 0 40,clip]{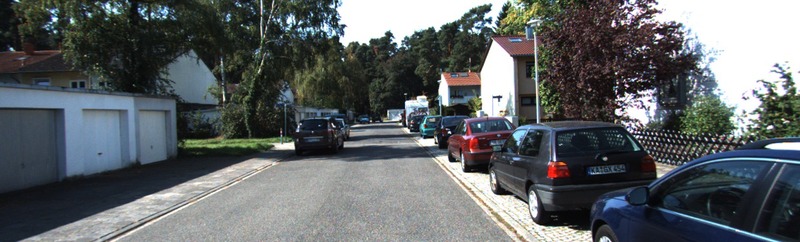}\\[-0.5mm]

 \includegraphics[width=0.32\textwidth,trim=0 0 0 40,clip]{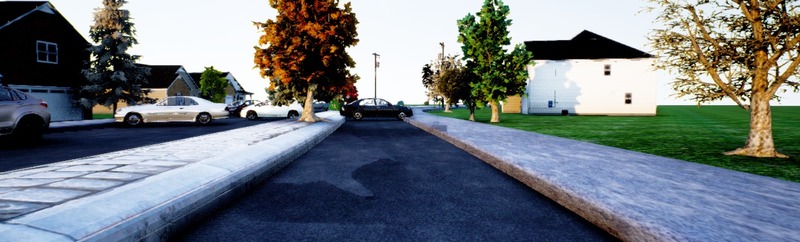} &
\includegraphics[width=0.32\textwidth,trim=0 0 0 40,clip]{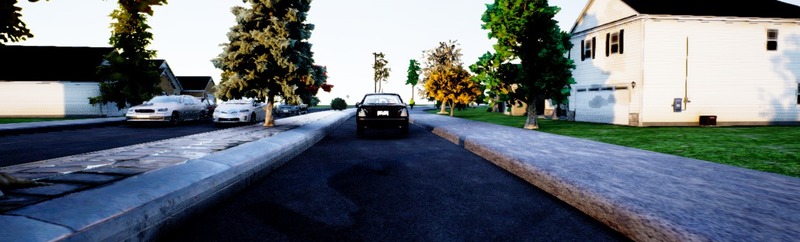} & 			\includegraphics[width=0.32\textwidth,trim=0 0 0 40,clip]{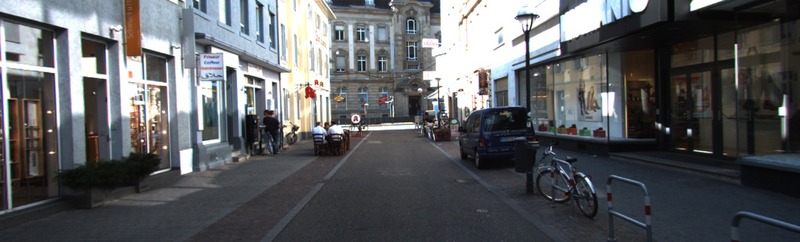}\\[-0.5mm]
						\includegraphics[width=0.32\textwidth,trim=0 0 0 40,clip]{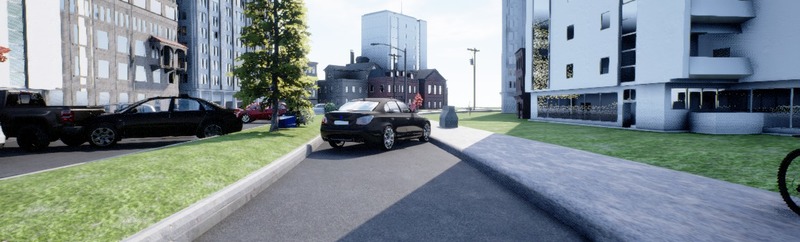} &
\includegraphics[width=0.32\textwidth,trim=0 0 0 40,clip]{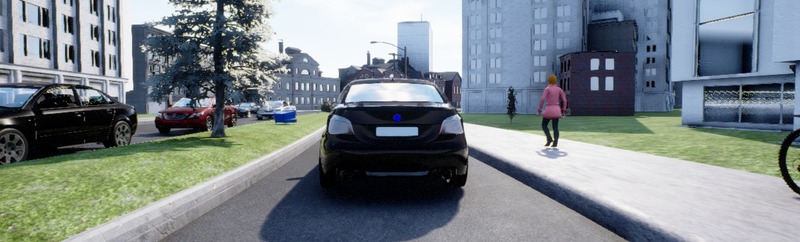} & 			\includegraphics[width=0.32\textwidth,trim=0 0 0 40,clip]{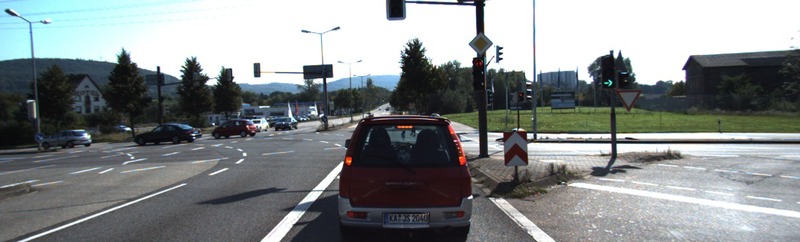}\\[-0.5mm]

\includegraphics[width=0.32\textwidth,trim=0 0 0 40,clip]{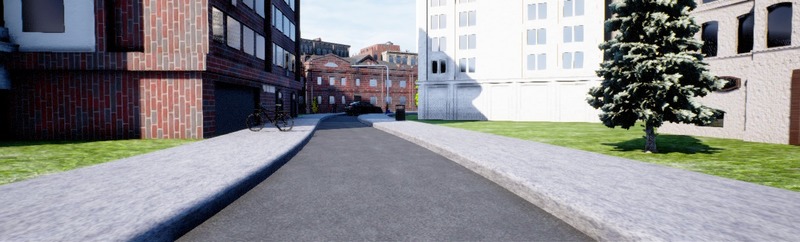} &
\includegraphics[width=0.32\textwidth,trim=0 0 0 40,clip]{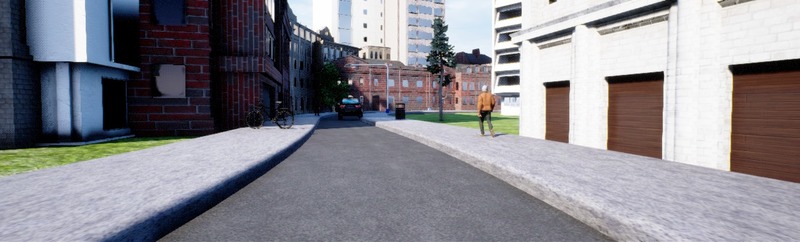} & 			\includegraphics[width=0.32\textwidth,trim=0 0 0 40,clip]{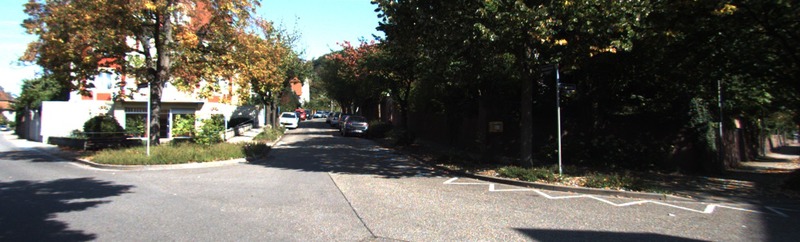}\\[-0.5mm]

\includegraphics[width=0.32\textwidth,trim=0 0 0 40,clip]{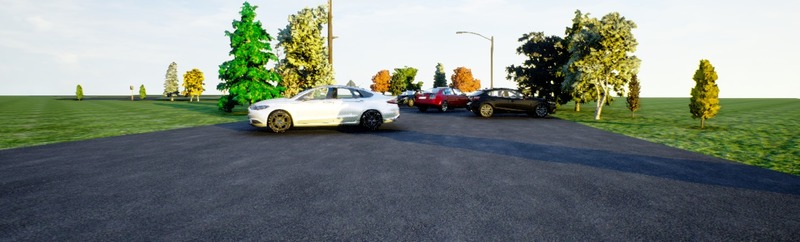} &
\includegraphics[width=0.32\textwidth,trim=0 0 0 40,clip]{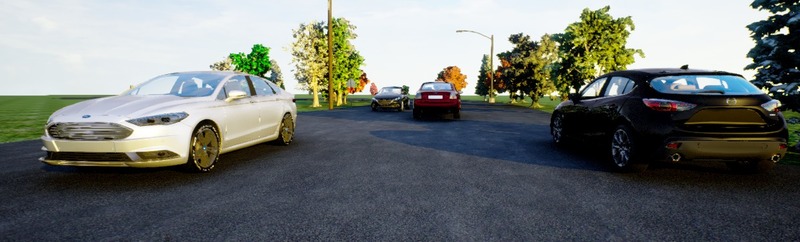} & 			\includegraphics[width=0.32\textwidth,trim=0 0 0 40,clip]{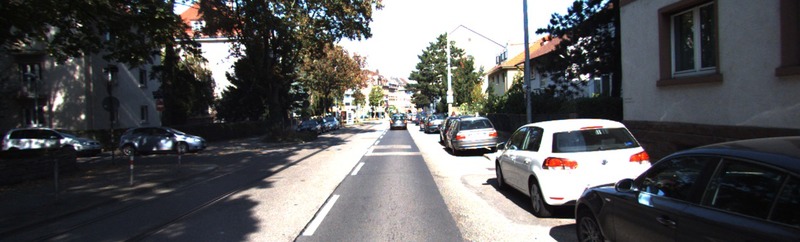}\\[-0.5mm]

\end{tabular}
    \caption{\footnotesize {\bf (left)} samples from our prob. grammar, {\bf (middle)} Meta-Sim's corresponding samples, {\bf (right)} random samples from KITTI}
\label{fig:metasim_sdr}
\vspace{-3mm}
\end{figure*}

\begin{figure*}
\begin{tabular}{c c}
  \animategraphics[loop,width=\columnwidth]{10}{figs/3d/gifs/car8-}{0}{28} &
    \animategraphics[loop,width=\columnwidth]{10}{figs/3d/gifs/car40-}{0}{28}
\end{tabular}
\caption{\footnotesize {Animation showing evolution of a scene through Meta-Sim training (animation works on Adobe Reader)}}
\label{fig:metasim_anim}
\vspace{-3mm}
\end{figure*}

\begin{figure*}[t!]
\vspace{-0mm}
    \centering
    \addtolength{\tabcolsep}{-4.6pt}
    \begin{tabular}{ccc}

			\includegraphics[width=0.32\linewidth]{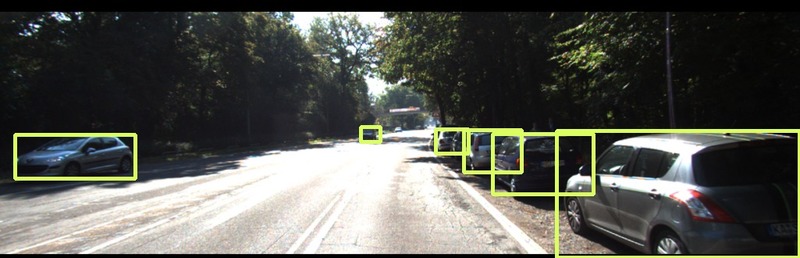} &
			\includegraphics[width=0.32\linewidth]{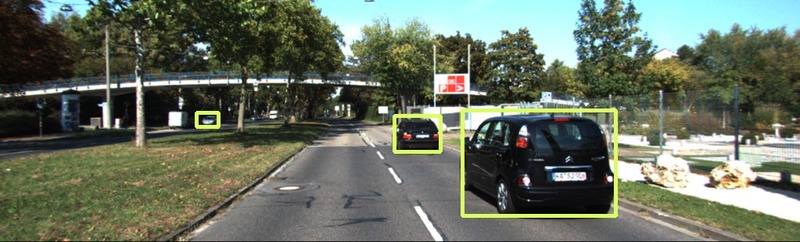} &
			\includegraphics[width=0.32\linewidth]{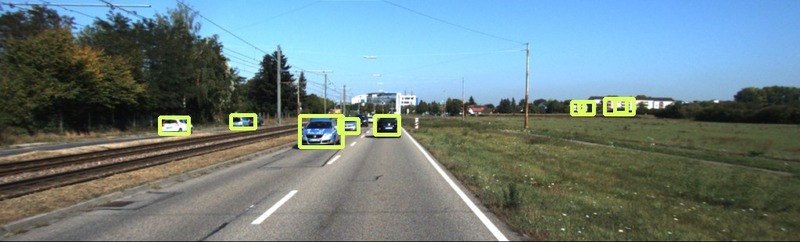} \\[-0.5mm]

			\includegraphics[width=0.32\linewidth]{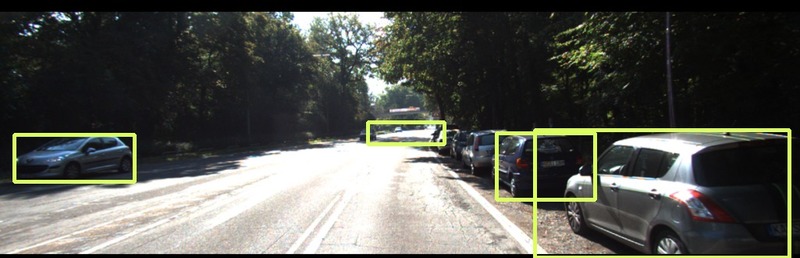} &
			\includegraphics[width=0.32\linewidth]{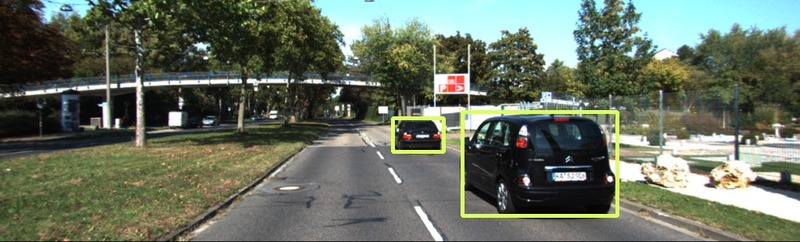} &
			\includegraphics[width=0.32\linewidth]{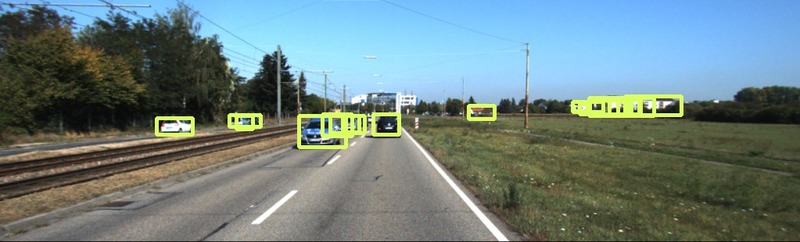} 
    \end{tabular}
    \caption{\footnotesize Car detection results {\bf (top)} of task network trained with Meta-Sim vs {\bf (bottom)} trained with our prob. grammar}
\label{fig:kitti_detection}
\vspace{-5.5mm}
\end{figure*}

Current open-source self driving simulators~\cite{Dosovitskiy17,AirSim} do not offer the amount of low level control on object attributes that we require in our model. We thus turn to probabilistic grammars for road scenarios~\cite{sdr18,synscapes}.   
Specifically, SDR~\cite{sdr18} is a road scene grammar that has been shown to outperform existing synthetic datasets as measured by sim-to-real performance. 
We adopt a simpler version of SDR and implement portions of their grammar as our probabilistic grammar. Specifically, we remove support for intersections and side-roads for computational reasons. The exact parameters of the grammar used can be found in the supplementary material.
We use the Unreal Engine 4 (UE4)~\cite{unreal} game engine for the 3D rendering from scene graphs.
Fig.~\ref{fig:metasim_sdr}(left column) shows example renderings of scenes generated using our version of the SDR grammar. The grammar parameters were mildly tuned, since we aim to have our model do the heavy lifting in subsequent parts.
\vspace{-4mm}
\paragraph{Task Network.} We use Mask-RCNN~\cite{he2017mask} with a Resnet-50-FPN backbone (ImageNet initialized) detection head as our task network for object detection.
\vspace{-4mm}
\paragraph{Experimental Setup.}
Following SDR~\cite{sdr18}, we use car detection as our task. Validation data $V$ is formed by taking 100 random images (and their labels) from the KITTI train set. The rest of the training data (images only) forms $X_R$. We report results on the KITTI val set. Training and finer details can be found in the supplementary material.

\vspace{-4mm}
\paragraph{Complexity.} 
To reduce training complexity (coming from rendering and numerical gradients), we train Meta-Sim to optimize specific parts of the scene sequentially. We first train to optimize attributes of cars. Next, we optimize car and camera parameters, and finally add parameters of context elements (buildings, pedestrians, trees) together to the training. Similarly, we decouple distribution and task training. We first train the above with MMD, and finally optimize all parameters above with the meta task loss.

\vspace{-4mm}
\paragraph{Quantitative Results.}
Table~\ref{tbl:3d} reports the average precision at 0.5 IoU of the task network trained using data generated from different methods, when tested on the KITTI val set. We see that training with Meta-Sim beats just using the data from the probabilistic grammar. 

\begin{table}[h!]
\vspace{-2mm}
\centering
\begin{small}
\begin{tabular}{c|c|c|c}
Data & Easy & Moderate & Hard \\
\hline\hline
Prob. Grammar &  63.7 & 63.7 & 62.2\\
MetaSim (Cars)  & 66.4 & \textbf{66.5} & 65.6\\
+ Camera  & 65.9 &  66.3 & 65.9\\
+ Context  & 65.9 &  66.3 & 66.0\\
+ Task Loss  &  \textbf{66.7} &  66.3 & \textbf{66.2}\\
\hline
\end{tabular}
\end{small}
\vspace{1mm}
\caption{AP @ 0.5 IOU for car detection on the KITTI val dataset}
\label{tbl:3d}
\vspace{-3mm}
\end{table}
Training the task network online with meta-sim and offline on final generated data results in similar final detection performance. This ensures the quality of the final generated data, since training while the transformation of data is being learned could be seen as data augmentation.
\vspace{-4mm}
\paragraph{Bridging the appearance gap.} We additionally add a state-of-the-art image-to-image translation network, MUNIT~\cite{munit} to attempt to bridge the appearance gap between the generated synthetic images and real images. Table~\ref{tbl:3d_munit} shows training with image-to-image translation still leaves a performance gap between MetaSim and the baseline, confirming our \emph{content gap} hypothesis.
\begin{table}[h!]
\vspace{-2mm}
\centering
\begin{small}
\begin{tabular}{c|c|c|c}
Data & Easy & Moderate & Hard \\
\hline\hline
Prob. Grammar &  71.1 &  \textbf{75.5} & 65.3\\
Meta-Sim &  \textbf{77.5} & 75.1 &  \textbf{68.2}\\
\hline
\end{tabular}
\end{small}
\vspace{1mm}
\caption{Effect of adding image-to-image translation to bridge the appearance gap in generated images}
\label{tbl:3d_munit}
\vspace{-4mm}
\end{table}
\vspace{-4.5mm}
\paragraph{Training on $V$.} Since we have access to some labelled training data, a valid baseline is to train the models on $V$ (100 images from KITTI train split). In Table.~\ref{tbl:3d_real} we show the effect of only training with $V$ and finetuning using  $V$.

\begin{table}[h!]
\vspace{-2mm}
\centering
\begin{small}
\begin{tabular}{c|c|c|c}
TaskNet Initialization & Easy & Moderate & Hard \\
\hline\hline
ImageNet & 61.2 & 62.0 & 60.7\\
Prob. Grammar & 71.3 & 72.7 & 72.7\\
Meta-Sim (Task Loss) & \textbf{72.4} & \textbf{73.9} & \textbf{73.9}\\
\hline
\end{tabular}
\end{small}
\vspace{1mm}
\caption{Effect of finetuning on $V$}
\label{tbl:3d_real}
\vspace{-4mm}
\end{table}

\vspace{-4mm}
\paragraph{Qualitative Results.}
Fig.~\ref{fig:metasim_sdr} shows a few outputs of Meta-Sim compared to the inputs sampled from the grammar, alongwith a few random samples from KITTI(train). There is a noticeable difference, as Meta-Sim's cars are well aligned with the road, and the distances between cars are meaningful. Also notice the small changes in camera, and the differences in the context elements, including houses, trees and pedestrians. The last row in Fig.~\ref{fig:metasim_sdr} represents a failure case where Meta-Sim is unable to clear up a dense initial scene, resulting in collided cars. Interestingly, Meta-Sim perfectly overlaps two cars in the same image such that a single car is visible from the camera (first car in front of camera). This behavior is seen multiple times, indicating that the model learns to cheat its way to good data. Fig.~\ref{fig:metasim_anim} shows an animation representing how a scene evolves through meta-sim training. Elements are moved to final configurations sequentially, following our training procedure. We remind the reader that these scene configurations are learned with only image/task level supervision. In Fig.~\ref{fig:kitti_detection}, we show results of training the task network on our grammar vs. training with Meta-Sim. We observe fewer false positives and negatives than the baseline. Meta-Sim shows better recall and GT overlap. Both models lose in precision, arguably because of not training for similar classes like Bus/Truck which would be negative examples.

%% file: conclusion.tex
\vspace{-2mm}
\section{Conclusion}
\vspace{-2mm}
We proposed Meta-Sim, an approach that generates synthetic data to match real content distributions while optimizing performance on downstream (real) tasks. Our model learns to transform sampled scenes from a probabilistic grammar so as to satisfy these objectives. Experiments on two toy and one real task showcased that Meta-Sim generates quantitatively better and noticeably higher quality samples than the baseline. We hope this opens a new exciting direction for simulation in the computer vision community. Like any other method, it has its limitations. It relies on obtaining valid scene structures from a grammar, and hence is still limited in the kinds of scenes it can model. Inferring rules of the grammar from real images, learning to generate structure of scenes and introducing multimodality in the model are intriguing avenues for future work.
\vspace{-3mm}
\paragraph{Acknowledgements:} The authors would like to thank Shaad Boochoon, Felipe Alves, Gavriel State, Jean-Francois Lafleche, Kevin Newkirk, Lou Rohan, Johnny Costello,  Dane Johnston and  Rev Lebaredian for their help and support throughout this project.

%% file: supplementary.tex

\section{Appendix}

\begin{figure}[t!]
    \centering
    \addtolength{\tabcolsep}{-4.6pt}
    \begin{tabular}{cccc}

			\includegraphics[width=0.24\linewidth]{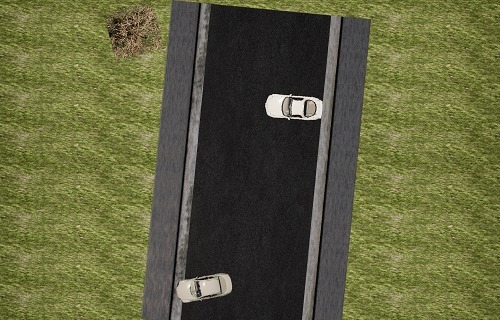} &
			\includegraphics[width=0.24\linewidth]{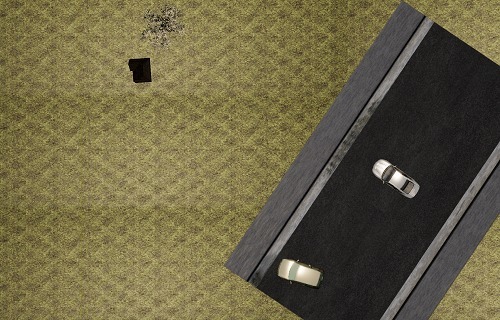} &
			\includegraphics[width=0.24\linewidth]{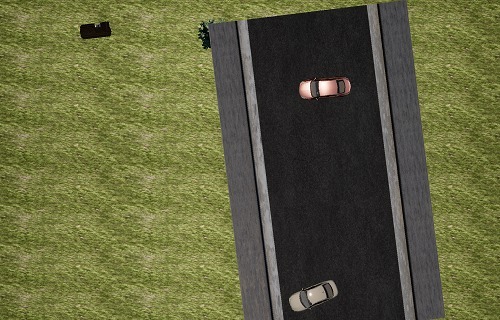} &
			\includegraphics[width=0.24\linewidth]{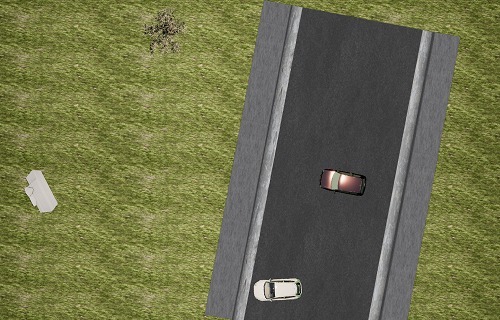} \\[-0.5mm]

		  \includegraphics[width=0.24\linewidth]{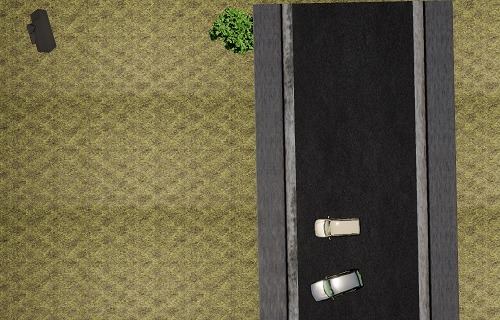} &
			\includegraphics[width=0.24\linewidth]{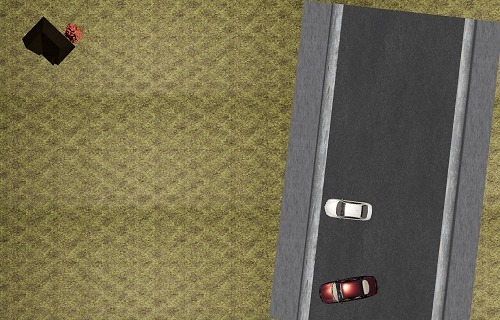} &
			\includegraphics[width=0.24\linewidth]{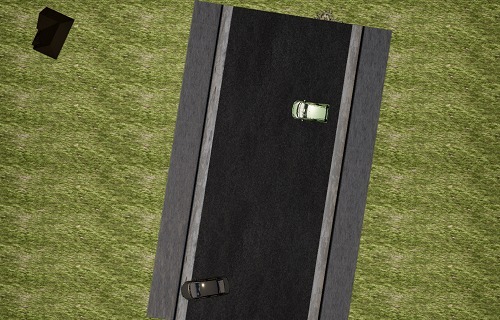} &
			\includegraphics[width=0.24\linewidth]{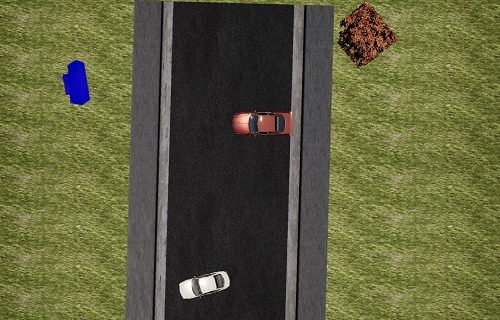} \\
			
			\includegraphics[width=0.24\linewidth]{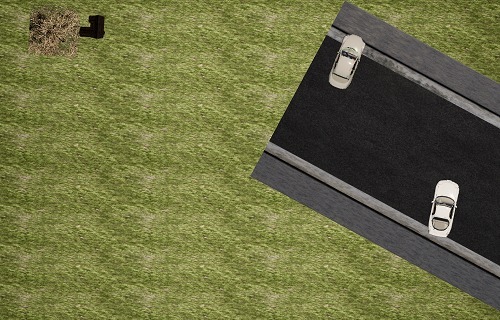} &
			\includegraphics[width=0.24\linewidth]{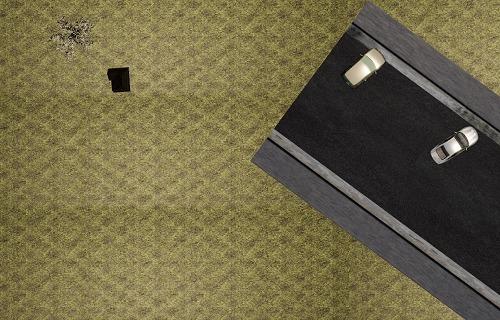} &
			\includegraphics[width=0.24\linewidth]{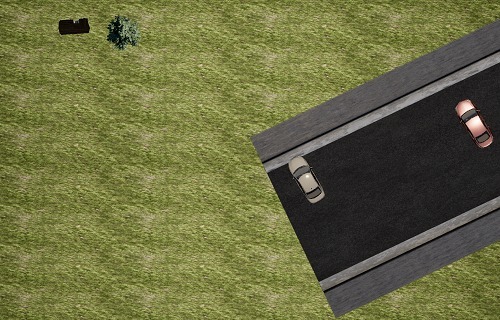} &
			\includegraphics[width=0.24\linewidth]{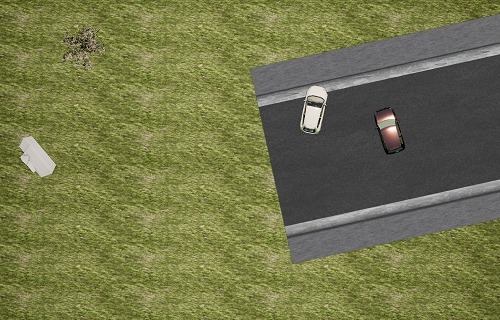} \\
			
			\includegraphics[width=0.24\linewidth]{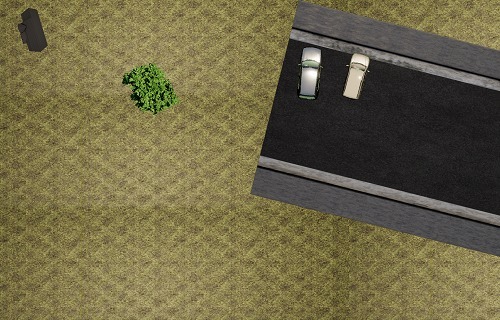} &
			\includegraphics[width=0.24\linewidth]{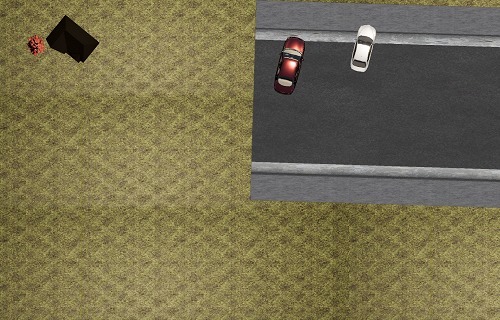} &
			\includegraphics[width=0.24\linewidth]{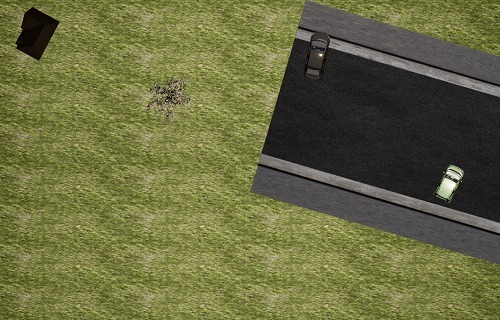} &
			\includegraphics[width=0.24\linewidth]{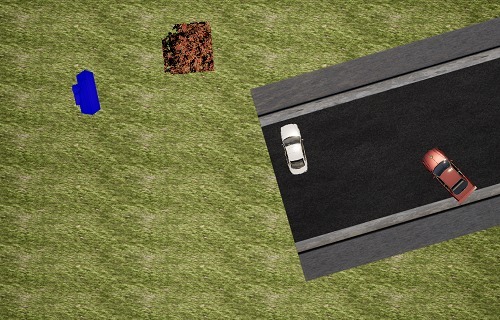} \\
			
    \end{tabular}
    \caption{\footnotesize Meta-Sim generated samples {\bf (top two rows)} and input scenes {\bf (bottom two rows)}}
\label{fig:2d_results}
\end{figure}

\begin{figure*}[t!]
\vspace{-3mm}
    \centering
    \begin{tabular}{c | c}
				
			\includegraphics[width=0.49\textwidth,trim=0 0 0 40,clip]{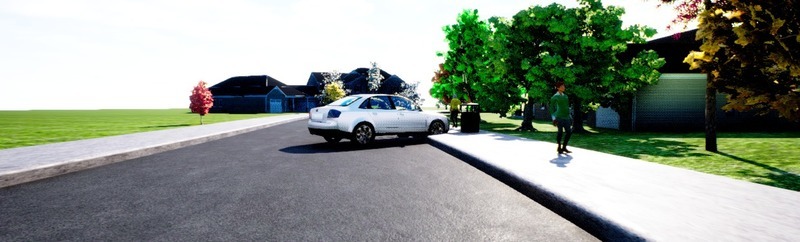} &
			\includegraphics[width=0.49\textwidth,trim=0 0 0 40,clip]{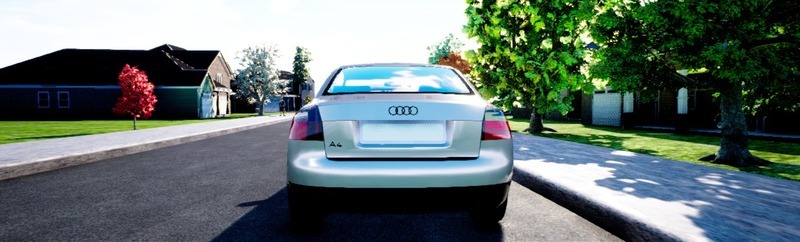} \\[-0.5mm]
			
				\includegraphics[width=0.49\textwidth,trim=0 0 0 40,clip]{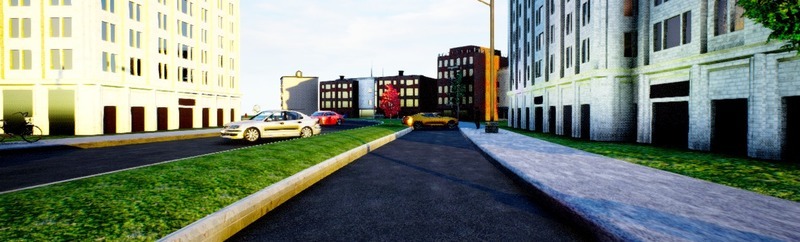} &
			\includegraphics[width=0.49\textwidth,trim=0 0 0 40,clip]{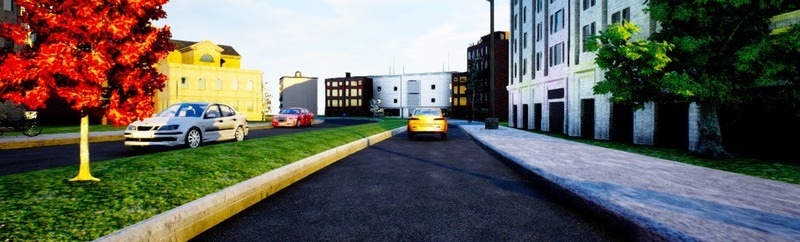} \\
			
				\includegraphics[width=0.49\textwidth,trim=0 0 0 40,clip]{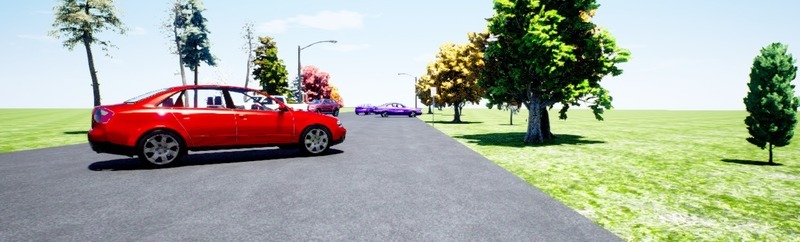} &
			\includegraphics[width=0.49\textwidth,trim=0 0 0 40,clip]{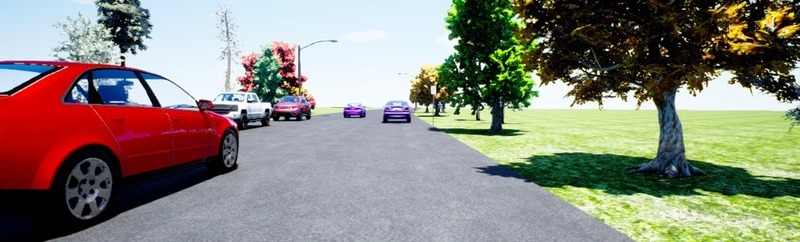} \\

			\includegraphics[width=0.49\textwidth,trim=0 0 0 40,clip]{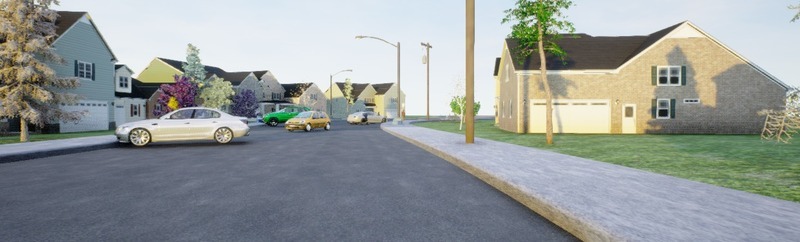} &
			\includegraphics[width=0.49\textwidth,trim=0 0 0 40,clip]{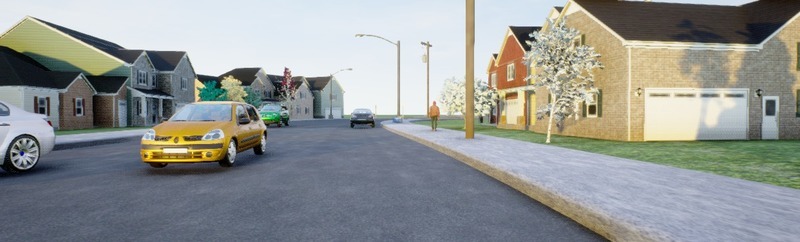} \\
			
				\includegraphics[width=0.49\textwidth,trim=0 0 0 40,clip]{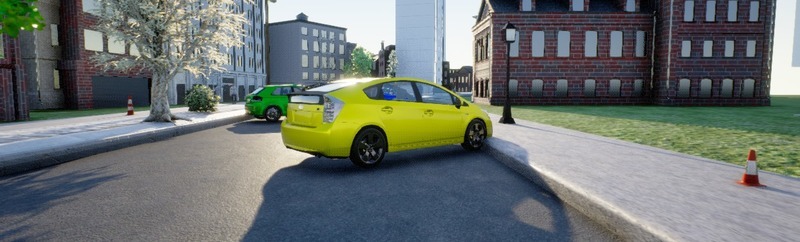} &
			\includegraphics[width=0.49\textwidth,trim=0 0 0 40,clip]{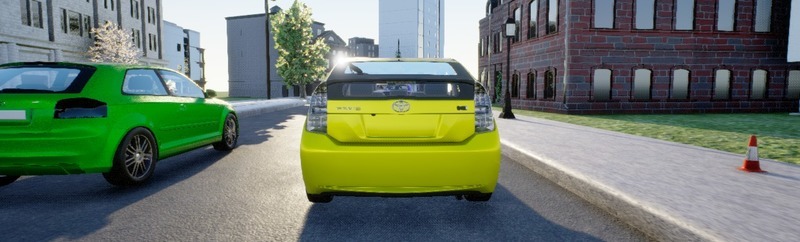} \\
			
				\includegraphics[width=0.49\textwidth,trim=0 0 0 40,clip]{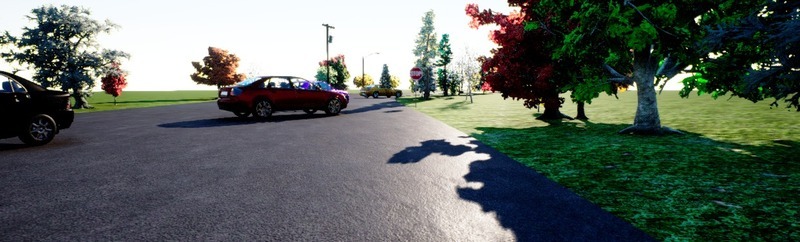} &
			\includegraphics[width=0.49\textwidth,trim=0 0 0 40,clip]{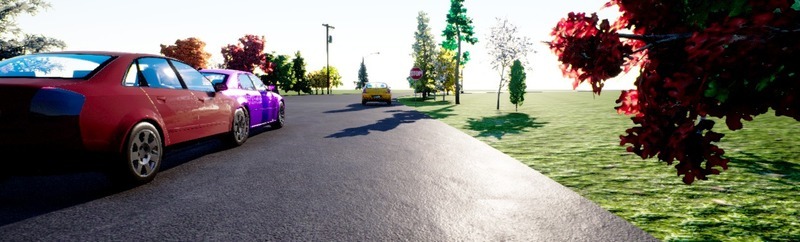} \\

			 \includegraphics[width=0.49\textwidth,trim=0 0 0 40,clip]{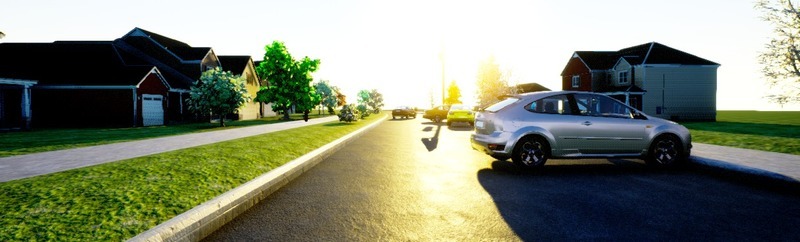} &
			\includegraphics[width=0.49\textwidth,trim=0 0 0 40,clip]{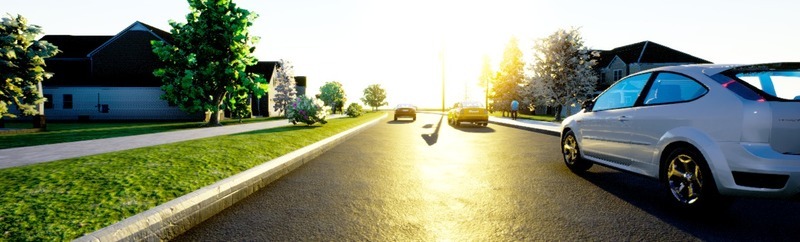} \\
			
			 \includegraphics[width=0.49\textwidth,trim=0 0 0 40,clip]{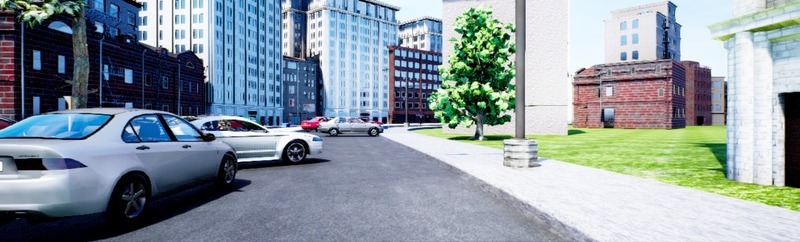} &
			\includegraphics[width=0.49\textwidth,trim=0 0 0 40,clip]{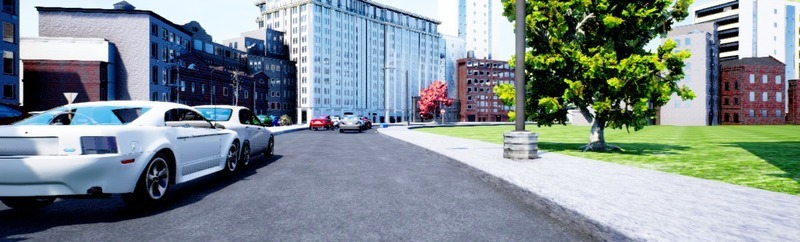} \\

    \end{tabular}
    \caption{\footnotesize {\textbf{Successful cases}: {\bf (left)} input scenes from the probabilistic} grammar, {\bf (right)} Meta-Sim's generated examples for the task of  car detection on KITTI. Notice how meta-sim learns to align objects in the scene, slightly change the camera position and move context elements such as buildings and trees, usually densifying the scene.}
\label{fig:metasim_sdr_success}
\vspace{-3mm}
\end{figure*}

\begin{figure*}[t!]
\vspace{-3mm}
    \centering
    \begin{tabular}{c | c}
				
			 \includegraphics[width=0.49\textwidth,trim=0 0 0 40,clip]{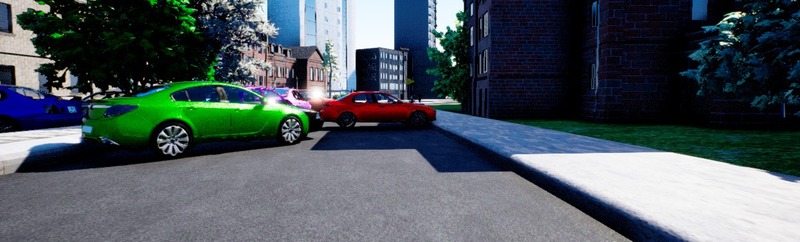} &
			\includegraphics[width=0.49\textwidth,trim=0 0 0 40,clip]{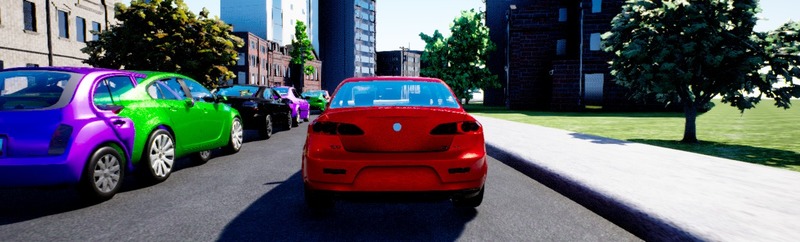} \\[-0.5mm]
			
				\includegraphics[width=0.49\textwidth,trim=0 0 0 40,clip]{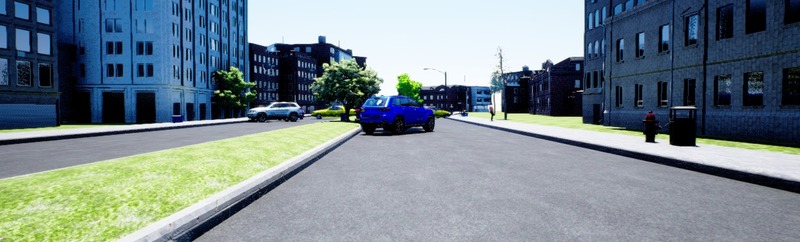} &
			\includegraphics[width=0.49\textwidth,trim=0 0 0 40,clip]{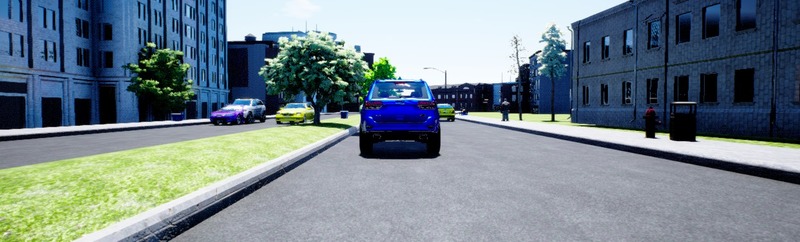} \\
			
				\includegraphics[width=0.49\textwidth,trim=0 0 0 40,clip]{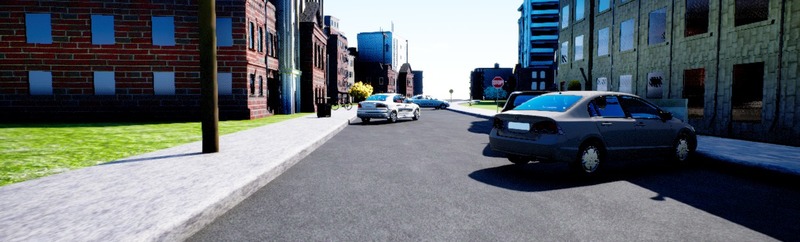} &
			\includegraphics[width=0.49\textwidth,trim=0 0 0 40,clip]{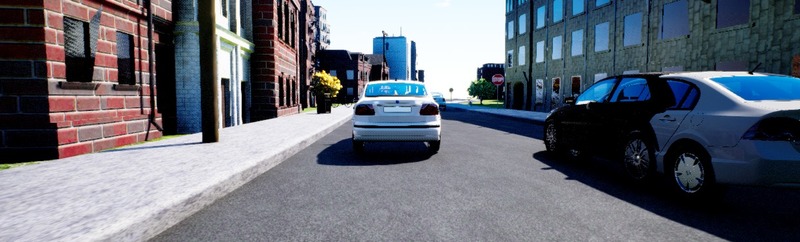} \\

			\includegraphics[width=0.49\textwidth,trim=0 0 0 40,clip]{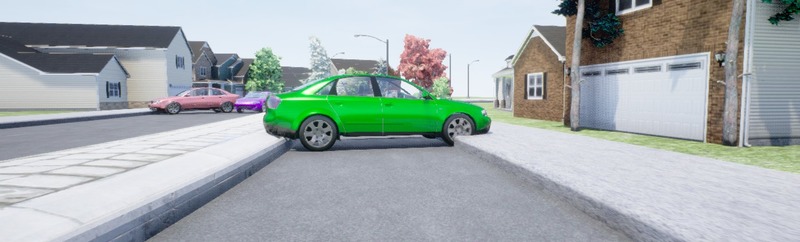} &
			\includegraphics[width=0.49\textwidth,trim=0 0 0 40,clip]{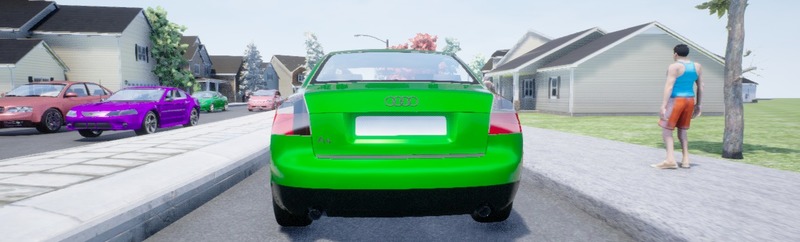} \\

    \end{tabular}
    \caption{ \footnotesize {\textbf{Failure cases}: \bf (left)} input scenes from the probabilistic grammar, {\bf (right)} Meta-Sim's generated examples for the task of car detection on KITTI. Initially dense scenes are sometimes unresolved, leading to collisions in the final scenes. There are unrealistic colours on cars since they are sampled from the prior and not optimized in this work. In the last row, meta-sim moves a car very close to the ego-car (camera). }
\label{fig:metasim_sdr_failure}
\vspace{-3mm}
\end{figure*}

\begin{figure*}[t!]
\vspace{-0mm}
    \centering
    \addtolength{\tabcolsep}{-4.6pt}
    \begin{tabular}{c | c}

\includegraphics[width=0.49\textwidth,trim=0 0 0 40,clip]{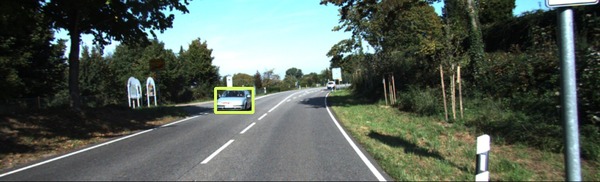} &
			\includegraphics[width=0.49\textwidth,trim=0 0 0 40,clip]{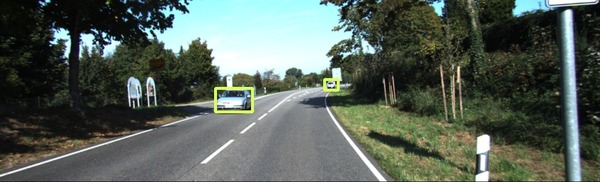} \\[-0.5mm]
			
				\includegraphics[width=0.49\textwidth,trim=0 0 0 40,clip]{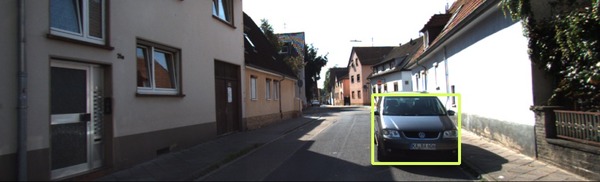} &
			\includegraphics[width=0.49\textwidth,trim=0 0 0 40,clip]{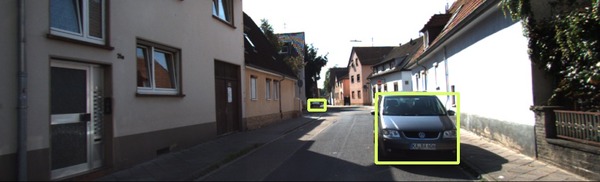} \\
			
				\includegraphics[width=0.49\textwidth,trim=0 0 0 40,clip]{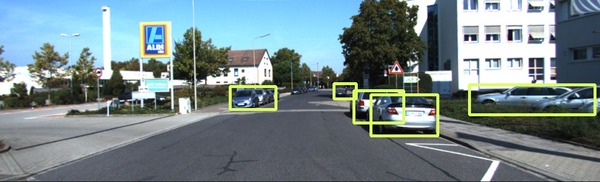} &
			\includegraphics[width=0.49\textwidth,trim=0 0 0 40,clip]{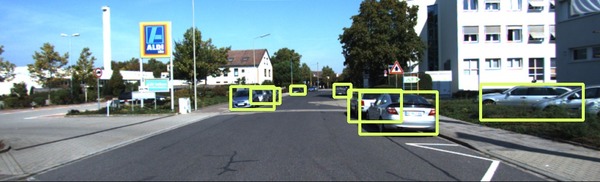} \\
				\includegraphics[width=0.49\textwidth,trim=0 0 0 40,clip]{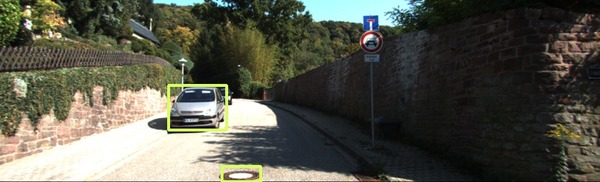} &
			\includegraphics[width=0.49\textwidth,trim=0 0 0 40,clip]{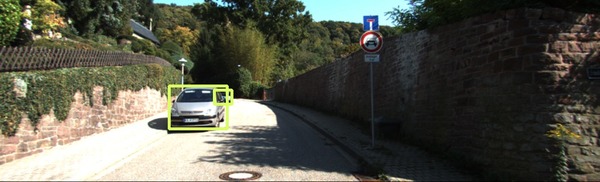} \\
			
				\includegraphics[width=0.49\textwidth,trim=0 0 0 40,clip]{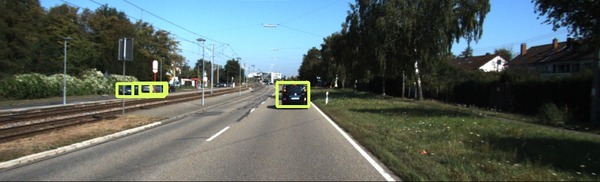} &
			\includegraphics[width=0.49\textwidth,trim=0 0 0 40,clip]{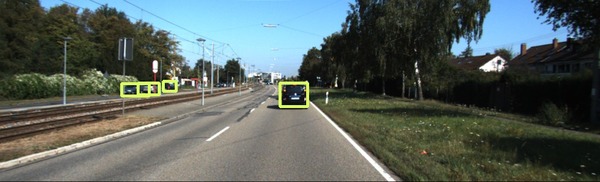} \\

			 \includegraphics[width=0.49\textwidth,trim=0 0 0 40,clip]{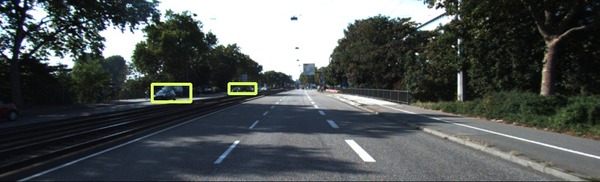} &
			\includegraphics[width=0.49\textwidth,trim=0 0 0 40,clip]{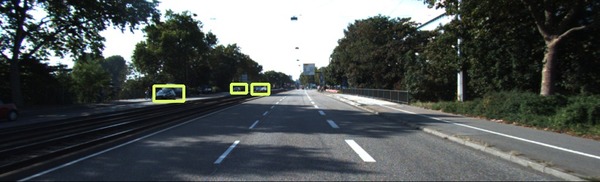} \\
			
			 \includegraphics[width=0.49\textwidth,trim=0 0 0 40,clip]{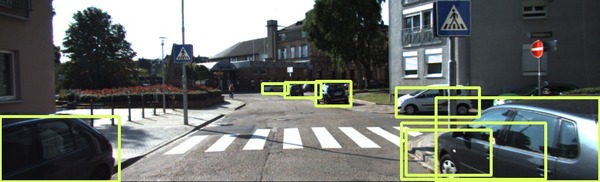} &
			\includegraphics[width=0.49\textwidth,trim=0 0 0 40,clip]{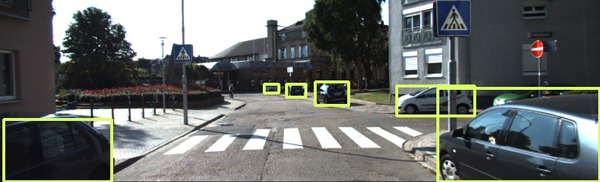} \\

    \end{tabular}
    \caption{\footnotesize Car detection results with Mask-RCNN trained  on {\bf (right)} the dataset generated by Meta-Sim, and {\bf (left)} the original dataset obtained by the probabilistic grammar. Notice how the models trained with meta-sim have lesser false positives and negatives.}
\label{fig:kitti_detection}
\vspace{-3mm}
\end{figure*}

We provide additional details about Meta-Sim, as well as include more results. 

\vspace{-2mm}
\subsection{Grammar and Attribute Sets}
First, we describe the probabilistic grammars for each experiment in more detail, as well as attribute sets. We remind the reader that the probabilistic grammar defines the structure of the scenes, and is used to sample scene graphs that are input to Meta-Sim. The Distribution Transformer transforms the attributes on the nodes in order to optimize its objectives. 

\vspace{-3mm}
\subsubsection{MNIST}
\label{ss:mnist}
\vspace{-2mm}
\paragraph{Grammar.} The sampling process for the probabilistic grammar used in the MNIST experiments is explained in the main paper, and we review it here. It samples a background texture and one digit texture (image) from the MNIST dataset~\cite{lecun1998mnist} (which has an equal probability for any digit), and then samples a rotation and location for the digit, to place it on the scene.

\vspace{-2mm}
\paragraph{Attribute Set.} The attribute set $s_A$ for each node consists of \emph{[class, rotation, locationX, locationY, size]}. Class is a one-hot vector, spanning all possible node classes in the graph. This includes, \emph{[scene, background, 0, ..., 9]}. \emph{scene} is the root node of the scene graph. The rotation, locations and size are floating point numbers between 0 to 1, and represent values in appropriate ranges \ie rotation in 0 to 360 degrees, each location within the parent's boundaries etc. 

\vspace{-4mm}
\subsubsection{Aerial Views (2D)}
\label{ss:aerial}
\vspace{-2mm}
\paragraph{Grammar.} The probabilistic grammar is explained in the main paper, and we summarize it again here. First, a background grass texture is sampled. Next, a (straight) road is sampled at
a location and rotation on the background. Two cars are then sampled with independent locations (constrained to be in the road by parametrizing in the road’s coordinate system), and rotations. In addition, a tree and a house are sampled and placed randomly on the scene. Each object in the scene also gets assigned a random texture from a repository of collected textures.
\vspace{-3mm}
\paragraph{Attribute Set.} The attribute set $s_{A}$ for each node in the Aerial 2D experiment consists of \emph{[class, rotation, locationX, locationY, size]}. Our choice of class includes road, car, tree, house and background, with class as a one-hot vector. Subclasses, such as type of tree, car, etc, are not included in the attribute set. They are sampled randomly in the grammar and are left unchanged by the Distribution Transformer. The values are normalized similarly to that in the MNIST experiment. We learn to transform the \emph{rotation} and \emph{location} for every object in the scene, but leave \emph{size} unchanged, \ie $size \not\in s_{A,mut}$. 

\vspace{-3mm}
\subsubsection{Driving Scenes (3D)}
\vspace{-2mm}
\paragraph{Grammar.} The grammar used in the driving experiment is adapted from ~\cite{sdr18} (Section III).
Our adaptation uses some fixed global parameters (\ie weather type, sky type) as compared to SDR. Our camera optimization experiment learns the height of the camera from the ground, while scene contrast, saturation and light direction are randomized. We also do not implement side-streets and parking lanes.

\vspace{-2mm}
\paragraph{Attribute Set. }
The attribute set $s_{A}$ for each node in the Driving Scene 3D experiment consists of \emph{[class, rotation, distance, offset]}. The classes, subclasses and ranges are treated exactly like in the Aerial 2D experiment. \emph{Distance} indicates how far along the parent spline the object is being placed and \emph{offset} indicates how much across the parent spline the object is placed. For each experiment, a subset of nodes in the graph is kept mutable. We optimized the attributes of cars, context elements such as buildings, foliage and people. Camera height from the global parameters was optimized, which was injected into the sky node encoded in the distance attribute. 

\vspace{-1mm}
\subsection{Training Details}
Let $a_{in}$ be the number of attributes in $s_A$ for each dataset. Our Distribution Transformer use Graph Convolutional layers with different weight matrices for each direction of the edge following~\cite{yao2018exploring}. We use 2 layers for the encoder and 2 layers for the decoder. In the MNIST and Aerial2D experiments, it has the following structure: Encoder ($a_{in} -> 16 -> 10$), and Decoder ($10 -> 16 -> a_{in}$). For 3D driving scenes, we use 28 and 18 feature size in the intermediate layers.  

After training the autoencoder step, the encoder is kept frozen and only the decoder is trained (\ie for the distribution matching and task optimization steps). For the MMD computation, we always use the pool1 and pool2 feature layers of the InceptionV3 network~\cite{szegedy2016rethinking}. For the MNIST and Aerial2D experiments, the images are not resized before passing them through the Inception network. For the Driving Scenes (3D) experiment, they are resized to 299 x 299 (standard inception input size) before computing the features. The task network training is iterative, our Distribution Transformer is trained for one epoch, followed by training of the task network for a few epochs.

\vspace{-5mm}
\subsubsection{MNIST} 
\vspace{-2mm}
For the MNIST experiment, 500 samples coming from the probabilistic grammar ared said to constitute 1 epoch. 
\vspace{-4mm}
\paragraph{Distribution Transformer.} We first train the autoencoder step of the Distribution Transformer for 8 epochs, with batch size of 8 and learning rate 0.001 using the adam optimizer~\cite{kingma2014adam}. Next, training is done with the joint loss, i.e. Distribution Loss and Task Loss, for 100 epochs. Note that when training with the task loss, there is only get 1 gradient step per epoch. We note that the MNIST tasks could be solved with the Distribution Loss alone, in around 3 epochs (since we can backpropagate the gradient for the distribution loss per minibatch when training only with the Distribution Loss).
\vspace{-4mm}
\paragraph{Task Network.} After getting each epoch of data from the Distribution Transformer, the Task Network is trained with this data for 2 epochs. The optimization was done using SGD with learning rate 0.01 and momentum 0.9. Note that the task network is not trained from scratch every time. Rather, training is resumed from the checkpoint obtained in the previous epoch. 
\vspace{-2mm}
\subsubsection{Aerial Views (2D)} 
For the Aerial Views (2D) experiment, 100 samples coming from the probabilistic grammar are considered to be 1 epoch.
\vspace{-4mm}
\paragraph{Distribution Transformer.} Here, the autoencoder step is trained for 300 epochs using the adam optimizer with batch size 8 and learning rate 0.01, and then for an additional 100 epochs with learning rate 0.001. Finally, the model is trained jointly with the distribution and task loss for 100 epochs with the same batch size and a learning rate of 0.001. 
\vspace{-6mm}
\paragraph{Task Network.} After getting each epoch of data from the Distribution Transformer, the Task Network is trained with this data for 8 epochs using SGD with a learning rate of 1e-8. Similar to the previous experiment, the task-network training is resumed instead of training from scratch.
\vspace{-4mm}
\subsubsection{Driving Scenes (3D)} 
\vspace{-2mm}
For Driving Scenes (3D), 256 samples coming from the prob. grammar are considered to be 1 epoch.
\vspace{-3mm}
\paragraph{Distribution Transformer. } In this case, the autoencoder step is trained for 100 epochs using the adam optimizer with a batch size of 16 and learning rate 0.005. Next, the model is trained with the distribution loss for 40 epochs with the same batch size and learning rate 0.001, finally stopping at the best validation (in our case, performance on the 100 images taken from the training set) performance. Finally, the model is trained with the task loss for 8 epochs with the same learning rate.
\vspace{-4mm}
\paragraph{Task Network.} For task network, we use Mask-RCNN~\cite{he2017mask} with a FPN and Resnet-101 backbone. We pre-train the task network (with freely available data sampled from the probabilistic grammar), saving training time by not having to train the detector from scratch. We also do not train the task network from scratch every epoch. Rather, we resume training from the checkpoint from the previous epoch. We also heavily parallelize the rendering computation by working on multiple UE4 workers simultaneously and utilizing their batched computation features.
\vspace{-2mm}

\subsection{Additional Results}
\subsubsection{Aerial Views (2D)}
\vspace{-2mm}
Fig.~\ref{fig:2d_results} shows additional qualitative results for the Aerial 2D experiment, comparing the original samples to those obtained via Meta-Sim.  The network learns to utilize the translation equivariance of convolutions and makes sufficient transformations to the objects for semantic segmentation to work on the validation data. Translating objects (such as cars) in the scene is still important, since our task-network for Aerial-2D has a maximum receptive field which is a quarter of the longest side of the image.

\vspace{-2mm}
\subsubsection{Driving Scenes (3D)}
Fig~\ref{fig:metasim_sdr_success} shows a few positive results for Meta-Sim. Our approach learns to rotate cars to look more like those in KITTI scenes and learns to place objects to avoid collisions between objects. The examples show that it can handle these even with a lot of objects present in the scene, while failing in some cases. It has also learnt to push cars out of the view of the camera when required. Sometimes, it positions two cars perfectly on top of each other, so that only one car is visible from the camera. This is in fact a viable solution since we solve for occlusion and truncation before generating ground truth bounding boxes. We also notice that the model modifies the camera height slightly, and moves context elements, usually densifying scenes. 

Fig.~\ref{fig:metasim_sdr_failure} shows failure cases for Meta-Sim. Sometimes, the model does not resolve occlusions/intersections correctly, resulting in final scenes with cars intersection each other. Sometimes it places an extra car between two cars (top row), attempts a partial merge of two cars (second row), causes collision (third row) and fails to rotate cars in the driving lane (final row). In this experiment, we did not model the size of objects in the features used in the Distribution Transformer. This could be a major reason behind this, since different assets (cars, houses) have different sizes and therefore placing them at the same location is not enough to ensure perfect occlusion between them. Meta-Sim sometimes fails to deal with situations where the probabilistic grammar samples objects densely around a single location (Fig.~\ref{fig:metasim_sdr_failure}). Another failure mode is when the model moves cars unrealistically close to the ego-car (camera) and when it overlaps buildings to create unrealistic composite buildings.
The right column of fig.~\ref{fig:kitti_detection} show additional detection results on KITTI when the task-network is trained with images generated by Meta-Sim. The left column shows results when trained with samples from the probabilistic grammar. In general, we see see fewer false positives and negatives with Meta-sim. 
The results are also significant given that the task network has not seen any real KITTI images during training.